\pdfoutput=1

\documentclass[11pt]{article}

\usepackage[final]{acl}
\usepackage{booktabs,tabularx, colortbl} 
\usepackage{multirow}
\usepackage{float}
\usepackage{amsfonts} 
\usepackage{bm}
\usepackage{mathtools}
\usepackage{amsmath}

\usepackage{times}
\usepackage{latexsym}

\usepackage[T1]{fontenc}

\usepackage[utf8]{inputenc}

\usepackage{microtype}

\usepackage{inconsolata}

\usepackage{graphicx}
\usepackage{float}
\usepackage{times}
\usepackage{latexsym}

\usepackage[T1]{fontenc}

\usepackage[utf8]{inputenc}
\usepackage{microtype}
\usepackage{float}
\usepackage{inconsolata}

\usepackage{amsmath,amsfonts,bm}

\def\eqref#1{equation~\ref{#1}}

\def\floor#1{\lfloor #1 \rfloor}
\def\1{\bm{1}}

\def\mK{{\bm{K}}}

\def\mQ{{\bm{Q}}}

\def\mV{{\bm{V}}}

\def\mX{{\bm{X}}}

\DeclareMathAlphabet{\mathsfit}{\encodingdefault}{\sfdefault}{m}{sl}
\SetMathAlphabet{\mathsfit}{bold}{\encodingdefault}{\sfdefault}{bx}{n}

\def\emX{{X}}

\usepackage{hyperref}
\usepackage{url}
\usepackage{multirow}
\usepackage{graphicx}
\usepackage{booktabs} 
\usepackage{subcaption}
\title{Accurate KV Cache Quantization with Outlier Tokens Tracing}

\author{Yi Su$^{1,2}$\thanks{\; Equal Contribution.}, Yuechi Zhou$^{1,2}$\footnotemark[1], Quantong Qiu$^{1,2}$, Juntao Li$^{1,2}$\thanks{\; Corresponding author.} \\
\textbf{Qingrong Xia$^3$, Ping Li$^3$, Xinyu Duan$^3$, Zhefeng Wang$^3$, Min Zhang$^{1,2}$} \\  
 $^{1}$School of Computer Science and Technology, Soochow University\\
$^{2}$Key Laboratory of Data Intelligence and Advanced Computing, Soochow University \\
 $^{3}$Huawei Cloud\\
 \texttt{yisunlp@outlook.com}; 
 \texttt{\{ljt,minzhang\}@suda.edu.cn} \\
 }

\begin{document}
\maketitle
\begin{abstract}
The impressive capabilities of Large Language Models (LLMs) come at the cost of substantial computational resources during deployment. While KV Cache can significantly reduce recomputation during inference, it also introduces additional memory overhead. KV Cache quantization presents a promising solution, striking a good balance between memory usage and accuracy.
Previous research has shown that the Keys are distributed by channel, while the Values are distributed by token. Consequently, the common practice is to apply channel-wise quantization to the Keys and token-wise quantization to the Values. However, our further investigation reveals that a small subset of unusual tokens exhibit unique characteristics that deviate from this pattern, which can substantially impact quantization accuracy.
To address this, we develop a simple yet effective method to identify these tokens accurately during the decoding process and exclude them from quantization as outlier tokens, significantly improving overall accuracy. Extensive experiments show that our method achieves significant accuracy improvements under 2-bit quantization and can deliver a 6.4 times reduction in memory usage and a 2.3 times increase in throughput\footnote{Code is available at https://github.com/yisunlp/OTT.}.
\end{abstract}



\section{Introduction}
Large Language Models (LLMs) have significantly impacted various industries due to their powerful capabilities \citep{gpt4,llama,llama2,llama3,mistral}. However, their auto-regressive nature makes the generation process slow. Although using KV Cache can reduce decoding complexity from $O(n^2)$ to $O(n)$ by storing the Keys and the Values computed during inference, it introduces substantial memory overhead. This overhead scales with sequence length, batch size, and hidden dim, often creating a memory bottleneck and placing considerable pressure on resources during deployment. As a result, optimizing KV Cache management to enhance resource utilization and improve model throughput remains a critical challenge.

KV Cache affects throughput in two primary ways. First, its memory usage limits the scalability of batch sizes, reducing parallelism during decoding, and thus lowering throughput. Second, attention computation is delayed while waiting for the KV Cache to be transferred from memory to the computation unit. As the size of the KV Cache grows, the transmission time increases, decreasing throughput. Existing approaches mainly address this issue by optimizing hardware scheduling \citep{deepspeedinference,flashattention,flexgen,vllm} and reducing the size of the KV Cache \citep{KIVI,KV_QUANT,GEAR,H20,Streamingllm}. In this paper, we focus on the latter approach, KV cache compression.

One method of reducing the size of the KV Cache is to reduce the number of values that need to be stored, which is related to the shape of the KV Cache: [$num\_layers$, $batch\_size$, $num\_heads$, $sequence\_length$, $head\_dim$]. There are various compression methods in each dimension, including layer-wise KV Cache sharing \citep{layer1,layer2,layer3,layer4}, prefix sharing \citep{batch1,batch2}, head-wise KV Cache sharing \citep{MQA,GQA}, token eviction \citep{Streamingllm,H20,eviction3}, and low-rank projection \citep{lora1,lora2,lora3}.

Another strategy for reducing the size of KV Cache is quantization. However, unlike weight quantization, KV Cache quantization poses unique challenges due to the uneven distribution of the Keys and Values \citep{GEAR}. To enhance quantization accuracy, various methods have been proposed, including using low-rank matrices to approximate the error before and after quantization \citep{GEAR}, smoothing Key distributions through specific mappings \citep{quant4,lora3}, channel-wise Key and token-wise Value asymmetric quantization \citep{KIVI,KV_QUANT}, non-uniform quantization \citep{KV_QUANT, llmint8}, mixed-precision quantization \citep{qaq}, and Block Floating Point (BFP) quantization \citep{bfp}. Among these methods, channel-wise Key and token-wise Value asymmetric quantization has garnered much attention for its high accuracy and tuning-free nature. This technique operates under the assumption that some channels of the Keys have huge magnitudes and that the distribution of the Keys within the same channel is relatively uniform.

However, our further exploration reveals that a few outlier tokens deviate from this assumption.
Their Keys have very small magnitudes in the outlier channels with large magnitudes, which greatly affects the accuracy of quantization.
Based on these observations, we propose KV Cache Quantization with Outlier Tokens Tracing (\textit{OTT}), a simple yet effective method that identifies these tokens and excludes them from the quantization process, thereby improving quantization accuracy.
With hardware-friendly implementation, \textit{OTT} achieves significant accuracy improvements under 2-bit quantization, resulting in a 6.4× reduction in memory usage and a 2.3× increase in throughput. 

Overall, our contributions are as follows:
\begin{itemize}
    \item We investigate the outlier channels of the KV Cache and identify that some outlier tokens deviate from the previous assumptions.
    \item We propose KV Cache Quantization with Outlier Tokens Tracing (\textit{OTT}), a simple yet effective method to identify and exclude these tokens during quantization, thus improving overall quantization accuracy.
    \item Our method achieves significant accuracy improvements under 2-bit quantization, yielding a 6.4× reduction in memory usage and a 2.3× increase in throughput.
\end{itemize}

\section{Background}

\paragraph{Implementation of KV Cache.} 
Transformer-based \citep{transformer} LLMs typically utilize KV cache to prevent the redundant calculation of the attention scores and accelerate auto-regressive decoding. The generation process of LLMs with KV cache is divided into the prefill phase and the decoding phase \citep{pd}.
Given a prompt \( X = \{x_0, x_1, \ldots, x_{n-1}\} \) and tensor \( \displaystyle \mX \in \mathbb{R}^{b \times n \times d} \) after embedding, where \( b \) is the batch size, \( n \) is the length of the prompt, and \( d \) represents the hidden size, we will briefly describe the calculation process of the attention block, and we omit the number of heads in the multi-head attention mechanism.

\textit{i)} During the prefill phase, the Keys \( \displaystyle \mK_{<n} \) and Values \( \displaystyle \mV_{<n} \) are computed and cached by transforming \( \displaystyle \mX \) through the Key and Value weight matrices \( \mathbf{W}_k, \mathbf{W}_v \in \mathbb{R}^{d \times d} \) of each layer, which can be formulated as:
\[
\displaystyle \mK_{<n} = \displaystyle \mX \mathbf{W}_k, \quad \displaystyle \mV_{<n} = \displaystyle \mX \mathbf{W}_v.
\]
\textit{ii)} During the decoding phase, only the Keys and Values of the new token \( x_n \) need to be calculated, which are then combined with the cached Keys and Values to compute the new attention scores and outputs. For the current input tensor \( \displaystyle \mX_n \in \mathbb{R}^{b \times 1 \times d} \), we update the KV cache as follows:
\[
\displaystyle \mK = \displaystyle \mK_{<n} \Vert \displaystyle \mK_n, \quad \displaystyle \mV = \displaystyle \mV_{<n} \Vert \displaystyle \mV_n,
\]
where \( \displaystyle \mK_n = \displaystyle \mX_n \mathbf{W}_k \) and \( \displaystyle \mV_n = \displaystyle \mX_n \mathbf{W}_v \). We calculate the new attention output \( \text{ATT} \) as follows:
\begin{equation}
\displaystyle \mQ_n = \displaystyle \mX_n \mathbf{W}_q, \quad \text{ATT} = \sigma \left(\frac{\displaystyle \mQ_n \displaystyle \mK^{\top}}{\sqrt{d_k}}\right) \mathbf{V},
\end{equation}
where \( \mathbf{W}_q \) is the query weight matrix, \( \sqrt{d_k} \) is the normalization factor, and $\sigma$ is the softmax function.

\paragraph{Necessity of compression.} 
While KV cache reduces the computational complexity from $O(n^2)$ to $O(n)$, it introduces substantial GPU memory overhead, particularly with long sequence lengths and large batch sizes. For example, in the case of LLaMA-3-8B \citep{llama3}, where the number of layers $n_{\text{layers}}$ is 32, the number of heads $h$ is 8, the head dimension $d$ is 512, the input length $l$ is 8192, and the batch size $b$ is 64, performing inference with fp16 precision (which uses 2 bytes per value) requires $4bhdln_{\text{layers}}$ bytes to store the KV cache—equivalent to 256GB of memory. Thus, effectively compressing the KV cache is crucial to reducing GPU memory usage.

\paragraph{Uniform Quantization. }
In this paper, we focus on compressing the KV cache by reducing the bit-width needed to represent cached tensors. A straightforward approach is Uniform Quantization ~\citep{uniformquant}, which maps continuous numerical data to a discrete domain. Specifically, to quantize a high-precision matrix (e.g., float32) $\displaystyle \mX$ to a matrix $\displaystyle \mX'$ with $b$-bit precision, we first determine the quantization step size $q$. Each element $\displaystyle \emX_{i,j} \in \displaystyle \mX$ can be quantized to $\textrm{Q}(\displaystyle \emX_{i,j})$ as follows: 
\begin{equation}
\begin{split}
\textrm{Q}(\displaystyle \emX_{i,j}) &= \lfloor (\displaystyle \emX_{i,j}-\displaystyle \mX_{min})/q\rfloor, \\
q &= (\displaystyle \mX_{max}-\displaystyle \mX_{min})/(2^b-1),
\end{split}
\label{eq:2}
\end{equation}

where $\floor \cdot$ is the rounding function. 

\paragraph{Group Quantization. }
However, Uniform Quantization does not fully exploit the distribution characteristics of the data, which can lead to significant quantization errors, especially when there are outliers. A more advanced technique is Group Quantization ~\citep{groupquant}, which divides the matrix into multiple groups, expecting the data within each group to share similar distribution characteristics. Unlike Uniform Quantization, Group Quantization allows each group to have different quantization parameters, such as step size. This flexibility enables the method to better adapt to the local characteristics of the data, thereby reducing quantization errors while maintaining a low bit-width.
The channel-wise Key quantization and token-wise Value quantization proposed by KIVI \citep{KIVI} is a type of Group Quantization.

\section{Method}
In this section, we propose KV Cache Quantization with Outlier Tokens Tracing (\textit{OTT}).
We start with a preliminary exploration of the Keys and Values before introducing our method.

\begin{figure*}[t]
    \centering
    \begin{subfigure}{0.26\textwidth}
        \centering
        \includegraphics[width=\linewidth]{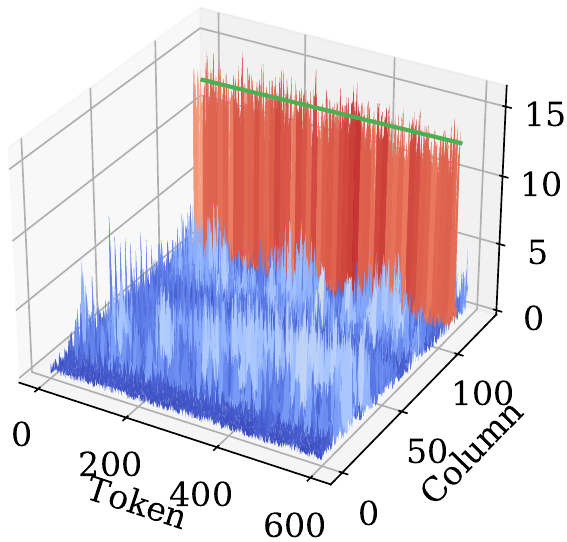}
        \caption{}
        \label{fig1:a}
    \end{subfigure}%
    \hspace{0.04\textwidth}
    \begin{subfigure}{0.26\textwidth}
        \centering
        \includegraphics[width=\linewidth]{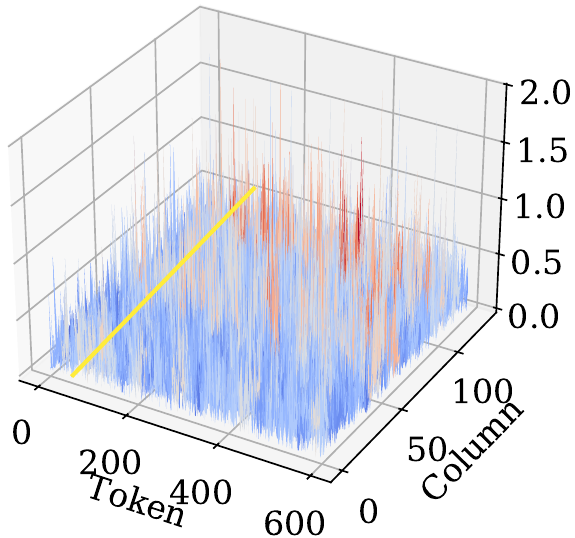}
        \caption{}
        \label{fig1:b}
    \end{subfigure}%
    \hspace{0.04\textwidth}
    \begin{subfigure}{0.31\textwidth}
        \centering
        \includegraphics[width=\linewidth]{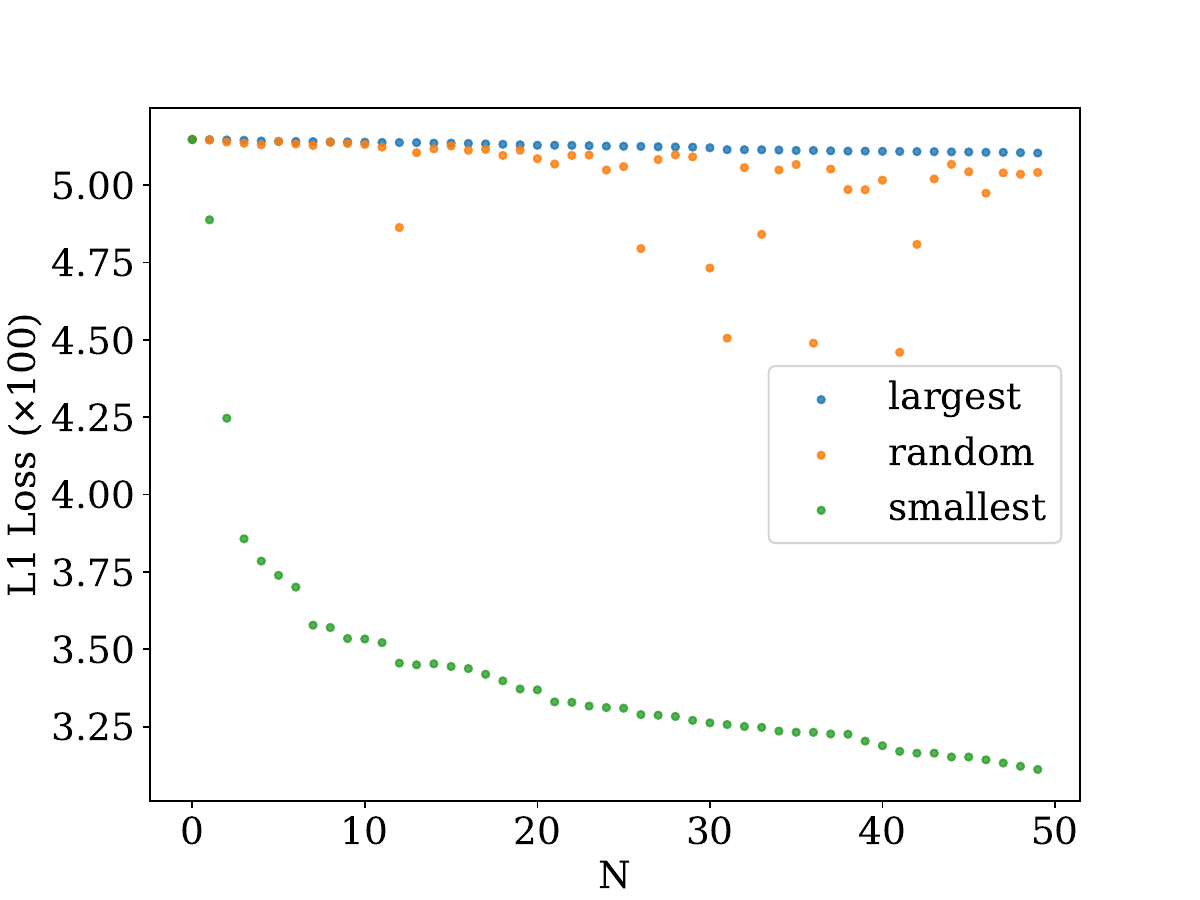}
        \caption{}
        \label{fig1:c}
    \end{subfigure}
  \begin{subfigure}{0.46\textwidth}
        \centering
        \includegraphics[width=\linewidth]{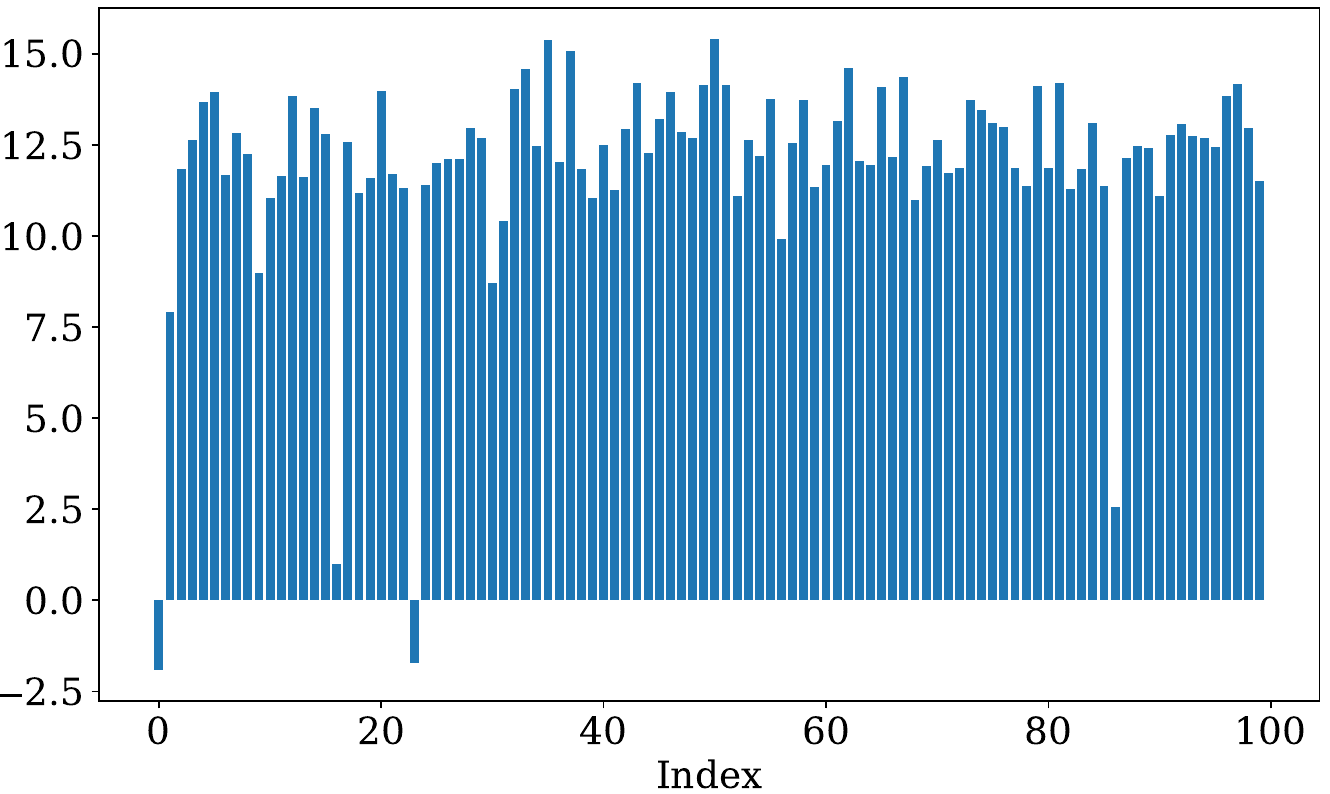}
        \caption{}
        \label{fig1:d}
    \end{subfigure}%
    \hspace{0.04\textwidth}
    \begin{subfigure}{0.46\textwidth}
        \centering
        \includegraphics[width=\linewidth]{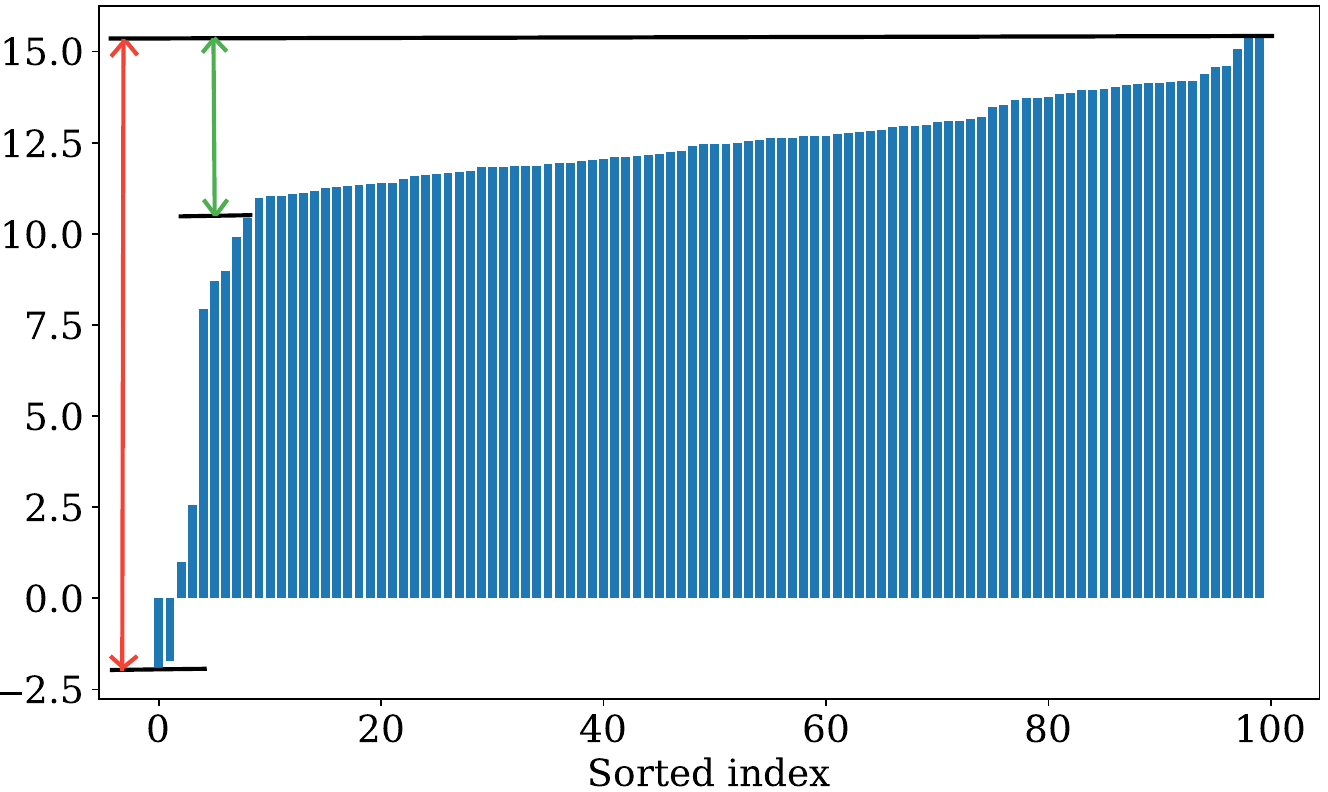}
        \caption{}
        \label{fig1:e}
    \end{subfigure}%
    
    \caption{
    Observations from preliminary experiments:
    (a) The Keys are distributed by channel and have some outlier channels.
    (b) The distribution of the Values does not exhibit any notable characteristics.
    (d) In certain outlier channels, a few tokens with low magnitude of Keys disrupt the originally uniform distribution within these channels.
    (e) Visualization of the sorted Keys in an outlier channel shows a rapid increase from a low value to very high values.
    (c) The L1 loss of attention output before and after quantization by retaining full-precision tokens based on different criteria. The best result is retaining full-precision tokens with the smallest magnitude of the Keys.
    }
    \label{fig1}
\end{figure*}

\subsection{Exploration of the Keys and Values.}
We conduct a series of preliminary experiments to gain a deeper understanding of the Keys and Values.
For illustration, we take a sentence generated by LLaMA-2-7b-chat-hf \footnote{\href{https://huggingface.co/meta-llama/Llama-2-7b-chat-hf}{https://huggingface.co/meta-llama/Llama-2-7b-chat-hf}} as an example.
Table \ref{tab:example} in the Appendix presents the prompt and the generated context.

\textbf{Distribution of the Keys and Values.} \
Figure \ref{fig1:a} and Figure \ref{fig1:b} display the magnitude of the Keys and Values from layer 10, head 17. Notably, some channels exhibit exceptionally large Keys, and within these channels, the distribution of the Keys appears relatively uniform. In contrast, the Values have no distinct characteristics. These observations are consistent with KIVI \citep{KIVI}.

\textbf{Distribution in outlier channels.} \
\label{outlier}
We further investigate the distribution of these outlier channels.
Figure \ref{fig1:d} shows the Keys in an outlier channel from layer 10, head 17 (we plot the first 100 tokens).
While the Keys generally exhibit a uniform distribution, a few tokens are notable exceptions. This pattern becomes clearer after sorting, as shown in Figure \ref{fig1:e}, where some Keys have very small values while others are significantly larger. These exceptions can substantially increase $\mX_{max}-\mX_{min}$ in Equation \ref{eq:2} during quantization, ultimately diminishing quantization accuracy.
Statistical analysis of these outliers can be found in Appendix \ref{statistical}.

\textbf{Identifying Outlier Tokens.} \
Intuitively, tokens with very small magnitude of the Keys in outlier channels are also likely to have smaller magnitude overall. To test this hypothesis, we plot the Keys from an outlier channel and the magnitude of the Keys across all channels (we plot the first 300 tokens). As shown in Figure \ref{fig:compare} in the Appendix, the results confirm our assumption, suggesting that we can efficiently and accurately identify these outlier tokens with the magnitude of the Keys.

\textbf{Removing Outlier Tokens.} \
\label{rule}
From our analysis, outlier tokens significantly impact the accuracy of quantization. By excluding these outlier tokens during quantization and retaining them with full precision, we can greatly reduce the loss of the attention output.
To investigate this further, we retain different numbers of tokens based on different selection criteria and compare the L1 loss of attention outputs before and after quantization. The results (Figure \ref{fig1:c}) reveal that retaining tokens with the largest keys yields the worst performance, while retaining those with the smallest Keys achieves the best results, aligning with our previous findings.

\subsection{\textit{OTT}: KV Cache Quantization with Outlier Tokens Tracing}
From the above observations, we find that some outlier tokens can seriously affect the accuracy of quantization.
So, we attempt to dynamically identify these tokens during the quantization process, exclude them during quantization, and retain their full-precision representations.
Our method consists of two components: quantization and decoding.

\textbf{Quantization} \ We define a fixed-size outlier pool with a capacity of $outlier\_num$ to store the Keys and Values of the outlier tokens.
Following KIVI \citep{KIVI}, we use channel-wise Key quantization and token-wise Value quantization.
We quantize KV Cache every $G$ (group size, a hyper-parameter in our method) steps.
So, at each quantization step, there are $G$ tokens to quant.
Based on the rule from Section \ref{rule}, we compute the magnitude of the Keys of each token as the criteria, and all the tokens (tokens to quant and tokens in the outlier pool) compete for a position in the outlier pool according to the criteria.
Once selected as an outlier token, the Keys and Values of the token are replaced with the mean values of all tokens to eliminate their impact on quantization.
When the outlier pool is full and replacements are needed, the tokens that are originally in the outlier pool but are defeated by the newly added tokens should return to their original positions.
But for the sake of simplicity, we retain an additional pool to store these tokens, and when this pool is full, we stop identifying outlier tokens.

\textbf{Decoding}  \ We maintain three types of KV Cache: the quantized KV Cache, the full-precision KV Cache, and the KV Cache stored in the outlier pool.
The full-precision KV Cache includes group tokens (when the group is not full, these tokens are not quantized and are kept in full precision) and recent tokens (a full-precision sliding window for the nearest tokens).
First, the Query is multiplied by all three types of Keys, and we concatenate the results to produce the attention scores. Next, we multiply these scores by their corresponding Values from each type and sum them to generate the final attention output. To enhance decoding efficiency, we utilize a CUDA fused kernel to multiply full-precision and quantized matrices.
We provide a simple mathematical formulation in Appendix \ref{math}.

\begin{figure*}
    \centering
    \includegraphics[width=1\linewidth]{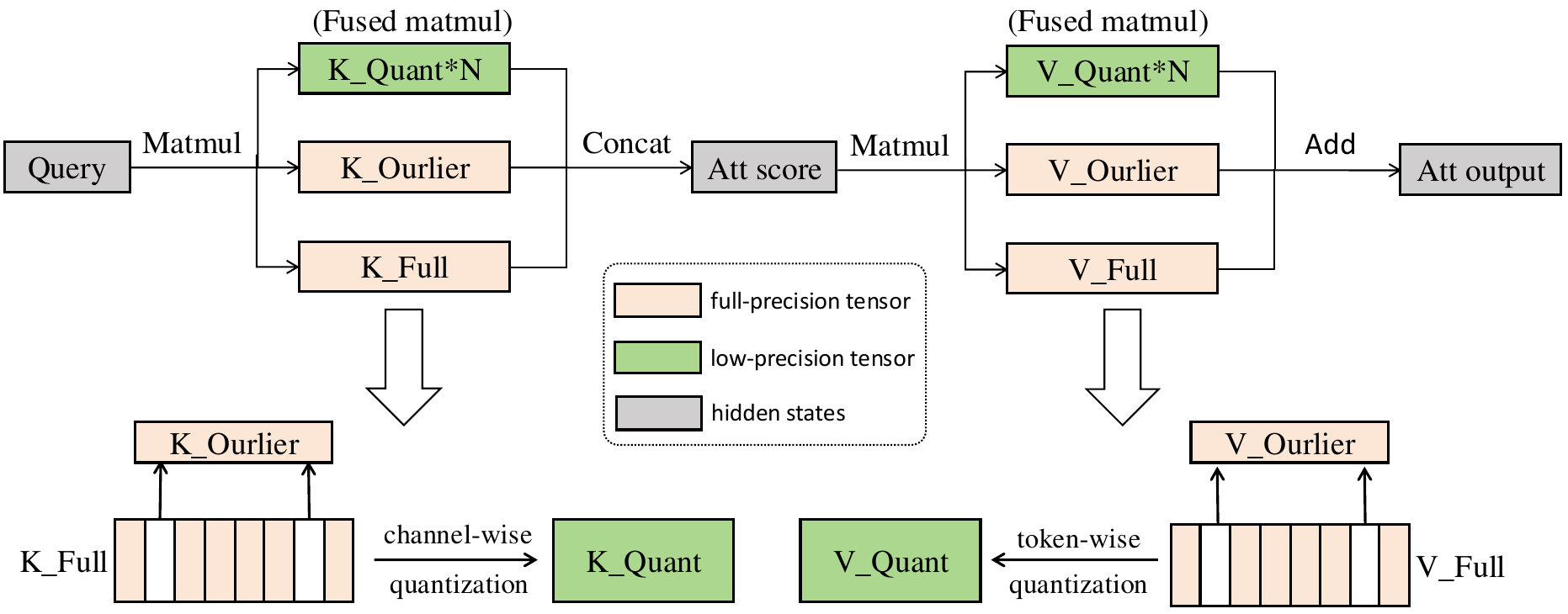}
    \caption{Overview of \textit{OTT}. Top: Decoding stage. Multiply the Query by each type of the Keys and concatenate the results to obtain the attention scores. Multiply the attention scores by each type of the Values and sum the results to get the attention output. Bottom: Quantization stage. Process the outlier tokens before quantization.}
    \label{fig:method}
\end{figure*}

\section{Experiments}

\subsection{Settings}

\textbf{Baselines and Models.}
KIVI \citep{KIVI} is currently one of the strongest tuning-free baseline with high compression efficiency and accuracy.
We use the same setting as KIVI, so we compare \textit{OTT} with KIVI and vanilla FP16 in our main experiments.
Due to differences in settings (e.g., compression frequency, compression factor, and the number of full-precision tokens), we do not include other KV Cache compression methods in the main experiment.
However, additional comparisons with these methods are provided in Appendix \ref{additional results}.
We use greedy decoding in our experiments.
We select two famous model families: LLaMA~\citep{llama2,llama3} and Mistral~\citep{mistral}. Specifically, we select LLaMA-2-7B-chat-hf, LLaMA-2-13B-chat-hf, LLaMA-3-8B-Instruct, and Mistral-7B-Instruct-v0.2. We also add experiments on LLaMA-2-70B-chat-hf in Appendix \ref{additional results}.

\textbf{Tasks.}
We evaluate our methods and the baselines on two benchmarks according to the length of input texts. For normal context length evaluation, we use arithmetic reasoning task Gsm8k ~\citep{gsm8k}, mainstream language and symbolic reasoning task BBH ~\citep{BBH}, and code completion task HumanEval ~\citep{humaneval}. For long context length evaluation, we choose four types of tasks in LongBench ~\citep{longbench} including Document QA, Summarization, Few-shot Learning and Code completion.
We provide more results of different benchmarks and baselines in Appendix \ref{additional baselines}.

\textbf{Details.}
We implement both KIVI and \textit{OTT} under 2-bit quantization. For KIVI, the group size (G) and residual length (R, size of the sliding window storing the nearest tokens) are set to 128. For \textit{OTT}, we use G = 128, R = 32, and set $outlier\_num$ to 3.
Notably, we set $outlier\_num$ to 0 for the first and second layers because we find that shallow layers have no outlier tokens (Ablation in Section \ref{sinknum}).
Regarding the additional pool used to store tokens evicted from the outlier pool, we find that a very small size is sufficient to retain all the eliminated tokens.
Therefore, we set its size to 32.
There are some differences in the processing of group and residual tokens between \textit{OTT} and KIVI.
KIVI compresses the Keys every $G$ steps, while compressing the Values at each step.
We choose to compress both the Keys and Values every $G$ steps for more consistent processing of KV Cache.
GSM8K and BBH are tested under LM Eval~\citep{lm_eval}. Humaneval follows the settings from InstructEval\footnote{\href{https://github.com/declare-lab/instruct-eval}{https://github.com/declare-lab/instruct-eval}}.
The experiments are conducted on NVIDIA A100 40G GPUs unless otherwise specified.

\subsection{Results}
\subsubsection{Normal context length evaluation}

Table \ref{tab:lm eval} presents the results of the normal context length evaluation across different models and methods. For Gsm8k and BBH, we report accuracy in the setting of few-shot, few-shot CoT, and zero-shot CoT. For HumanEval, we report pass@1 and pass@10 in the zero-shot setting.
The results illustrate that our method significantly outperforms KIVI across all settings. Notably, on BBH (3-CoT, LLaMA-3-8B-Instruct), \textit{OTT} achieves a 12.93\% improvement over KIVI. Compared to FP16, \textit{OTT} incurs minor accuracy loss in most settings. The largest accuracy drop occurs on BBH (3-CoT, Mistral-7B-Instruct), likely due to the high complexity of the task and the long generation length required.
Overall, \textit{OTT} can achieve significant performance improvements over KIVI.

\begin{table*}[t]

\centering
\small
\renewcommand{\arraystretch}{1.6} 
\setlength{\tabcolsep}{4.2pt} 
\begin{tabular}{lllllllllllll}
\toprule

\multicolumn{1}{l}{\multirow{2}{*}{\textbf{Dataset}}} & \multicolumn{3}{c}{\textbf{LLaMA-2-7B-chat-hf}} & \multicolumn{3}{c}{\textbf{LLaMA-2-13B-chat-hf}} & \multicolumn{3}{c}{\textbf{LLaMA-3-8B-Instruct}} & \multicolumn{3}{c}{\textbf{Mistral-7B-Instruct}}\\
\cmidrule(lr){2-4} \cmidrule(lr){5-7} \cmidrule(lr){8-10} \cmidrule(lr){11-13}
& \multicolumn{1}{c}{Fp16} & \multicolumn{1}{c}{KIVI} & \multicolumn{1}{c}{Ours}
& \multicolumn{1}{c}{Fp16} & \multicolumn{1}{c}{KIVI} & \multicolumn{1}{c}{Ours}
& \multicolumn{1}{c}{Fp16} & \multicolumn{1}{c}{KIVI} & \multicolumn{1}{c}{Ours}
& \multicolumn{1}{c}{Fp16} & \multicolumn{1}{c}{KIVI} & \multicolumn{1}{c}{Ours}
\\ \hline

Gsm8k (8) & 21.99	&16.30	&\textbf{21.38}	&36.54	&28.51	&\textbf{36.09}	&74.91	&63.15	&\textbf{72.55}	&42.91	&37.38	&\textbf{41.17} \\
+ CoT     & 21.30	&17.51	&\textbf{18.20}	&37.00	&31.77	&\textbf{36.92}	&76.72	&66.79	&\textbf{75.06}	&42.99	&37.45	&\textbf{41.39} \\
+ 0-CoT   & 24.11	&21.61	&\textbf{22.59}	&32.60	&29.19	&\textbf{31.31}	&40.64	&37.54	&\textbf{42.68}	&40.18	&33.81	&\textbf{37.98} \\
\hline

BBH (3)   & 33.34	&32.48	&\textbf{33.36}	&37.61	&36.20	&\textbf{37.43}	&45.77	&44.19	&\textbf{45.60}	&42.10	&40.29	&\textbf{42.02} \\
+ CoT     & 40.21	&34.00	&\textbf{35.17}	&47.38	&41.02	&\textbf{44.37}	&68.18	&47.38	&\textbf{60.31}	&51.33	&36.42	&\textbf{41.93} \\
+ 0-CoT   & 35.00	&33.30	&\textbf{34.25}	&35.86	&33.57	&\textbf{34.80}	&51.37	&44.19	&\textbf{48.89}	&41.74	&37.83	&\textbf{40.19} \\ 
\hline


HE (p@1) &12.19 &9.75 &\textbf{11.58} &7.92 &7.31 &\textbf{7.92} &40.24 &28.05 &\textbf{40.85} &40
.24 &32.92 & \textbf{35.36} \\
HE (p@10) &17.07 &12.19 &\textbf{14.63} &13.41 &11.58 &\textbf{15.24} &69.51 &56.09 &\textbf{67.68} &54.87 &50.00 & \textbf{54.26}  \\
\hline

Average       & 25.65	&22.14	&\textbf{23.90}	&31.04	&28.14	&\textbf{30.51}	&58.42	&48.16	&\textbf{56.70}	&44.55	&38.26	&\textbf{41.79} \\
\bottomrule
\end{tabular}
\caption{Results on GSM8K, BBH, and HumanEval (HE). \textbf{Bold}: the best results. We report accuracy for Gsm8k, BBH and Pass@k for HumanEval. Pass@k (p@k) refers to running each test question k times and calculating the average pass rate of the generated code. \textit{OTT} outperforms KIVI across all tasks, achieving the best results.} 
\label{tab:lm eval} 
\end{table*}

\subsubsection{Long context length tasks evaluation}
The main results of long context length evaluation are in table \ref{tab:longbench}. Our method outperforms KIVI in most settings, with only a tiny performance gap compared to the FP16 baseline. While KIVI maintains good accuracy on most tasks, it occasionally experiences significant performance drops (e.g., LLaMA-3-8B-Instruct, LCC: 56.58\% $\rightarrow$ 44.42\%).
However, \textit{OTT} does not encounter this situation, which suggests that our method achieves higher quantization accuracy than KIVI.

\begin{table*}[t]
\centering
\small
\renewcommand{\arraystretch}{1.4} 
\setlength{\tabcolsep}{3.0pt} 
\resizebox{2\columnwidth}{!}{%
\begin{tabular}{llccccccccc}
\toprule

\multicolumn{2}{l}{\multirow{1}{*}{\textbf{Model}}} & \multicolumn{1}{c}{\textbf{Qasper}} & \multicolumn{1}{c}{\textbf{GovReport}} & \multicolumn{1}{c}{\textbf{MultiNews}} & \multicolumn{1}{c}{\textbf{TREC}} & \multicolumn{1}{c}{\textbf{TriviaQA}} & \multicolumn{1}{c}{\textbf{SamSum}} & \multicolumn{1}{c}{\textbf{LCC}} & \multicolumn{1}{c}{\textbf{RepoBench-P}} & \multicolumn{1}{c}{\textbf{Avg}}  \\
\hline

\multirow{3}{*}{LLaMA-2-7B-chat-hf} & Fp16 & 20.04  &25.08  &23.02  &59.67  &85.39  &39.28  &59.59  &48.04  &45.01 \\
                           & KIVI & \textbf{20.43}	&19.97	&19.82	&59.67	&\textbf{85.16}	&37.70	&58.73	&47.24	&43.59 \\
                           & Ours & 19.95	&\textbf{21.56}	&\textbf{20.81}	&\textbf{59.67}	&85.00	&\textbf{39.10}	&\textbf{59.44}	&\textbf{48.51}	&\textbf{44.26} \\
\hline

\multirow{3}{*}{LLaMA-2-13B-chat-hf} & Fp16 & 17.42	&25.65	&23.35	&64.00	&86.52	&40.49	&49.80	&47.13	&44.30 \\
                            & KIVI & \textbf{20.10}	&20.65	&21.10	&63.67	&86.39	&39.51	&49.10	&43.95	&43.06 \\
                            & Ours & 18.81	&\textbf{22.29}	&\textbf{21.69}	&\textbf{64.00}	&\textbf{86.81}	&\textbf{40.35}	&\textbf{51.14}	&\textbf{47.71}	&\textbf{44.10} \\
\hline

\multirow{3}{*}{LLaMA-3-8B-Instruct} & Fp16 & 37.54  &31.04   &25.58 &69.67  &89.85  &40.50  &56.58  &51.01  &50.22 \\
                           & KIVI & 34.88	&28.43	&24.78	&69.33	&89.57	&40.09	&44.42	&45.54	&47.13 \\
                           & Ours & \textbf{36.75}	&\textbf{30.74}	&\textbf{24.94}	&\textbf{69.67}	&\textbf{89.74}	&\textbf{40.39}	&\textbf{52.37}	&\textbf{48.82}	&\textbf{49.18} \\
\hline

\multirow{3}{*}{Mistral-7B-Instruct-v0.2}& Fp16 & 24.35  &33.05  &25.77  &67.00  &86.84  &40.95  &57.24  &49.84  &48.13 \\
                           & KIVI & \textbf{24.20}	&30.98	&25.10	&66.33	&85.40	&41.05	&55.70	&48.18	&47.12 \\
                           & Ours & 23.78	&\textbf{31.37}	&\textbf{25.35}	&\textbf{66.33}	&\textbf{86.18}	&\textbf{41.25}	&\textbf{55.89}	&\textbf{48.32}	&\textbf{47.31} \\

\bottomrule
\end{tabular}
}
\caption{Main results on LongBench. We report accuracy for TREC, Rouge-L for GovReport and SamSum, edit similarity (Levenshtein distance \citep{similarity}) for LCC and RepoBenchP, and F1 score for the other tasks. \textbf{Bold}: the best results for each setting. \textit{OTT} demonstrates superior performance on average.} 
\label{tab:longbench} 
\end{table*}

\subsection{Efficiency comparison}

\begin{figure*}[t]

    \centering
    \begin{subfigure}[b]{0.32\textwidth} 
        \includegraphics[width=\textwidth]{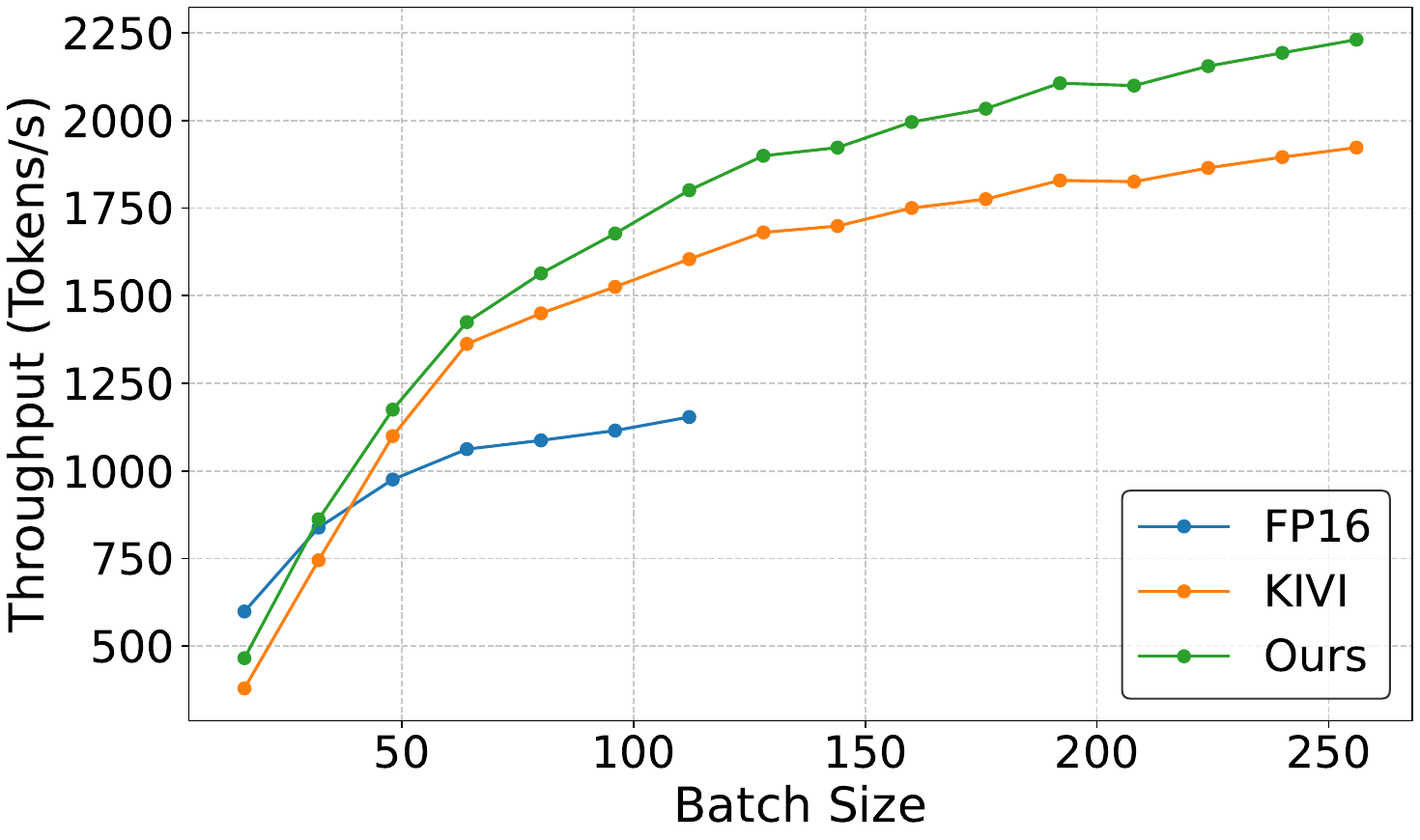}
        \caption{System throughput}
        \label{fig: throughput}
    \end{subfigure}
    \begin{subfigure}[b]{0.32\textwidth} 
        \includegraphics[width=\textwidth]{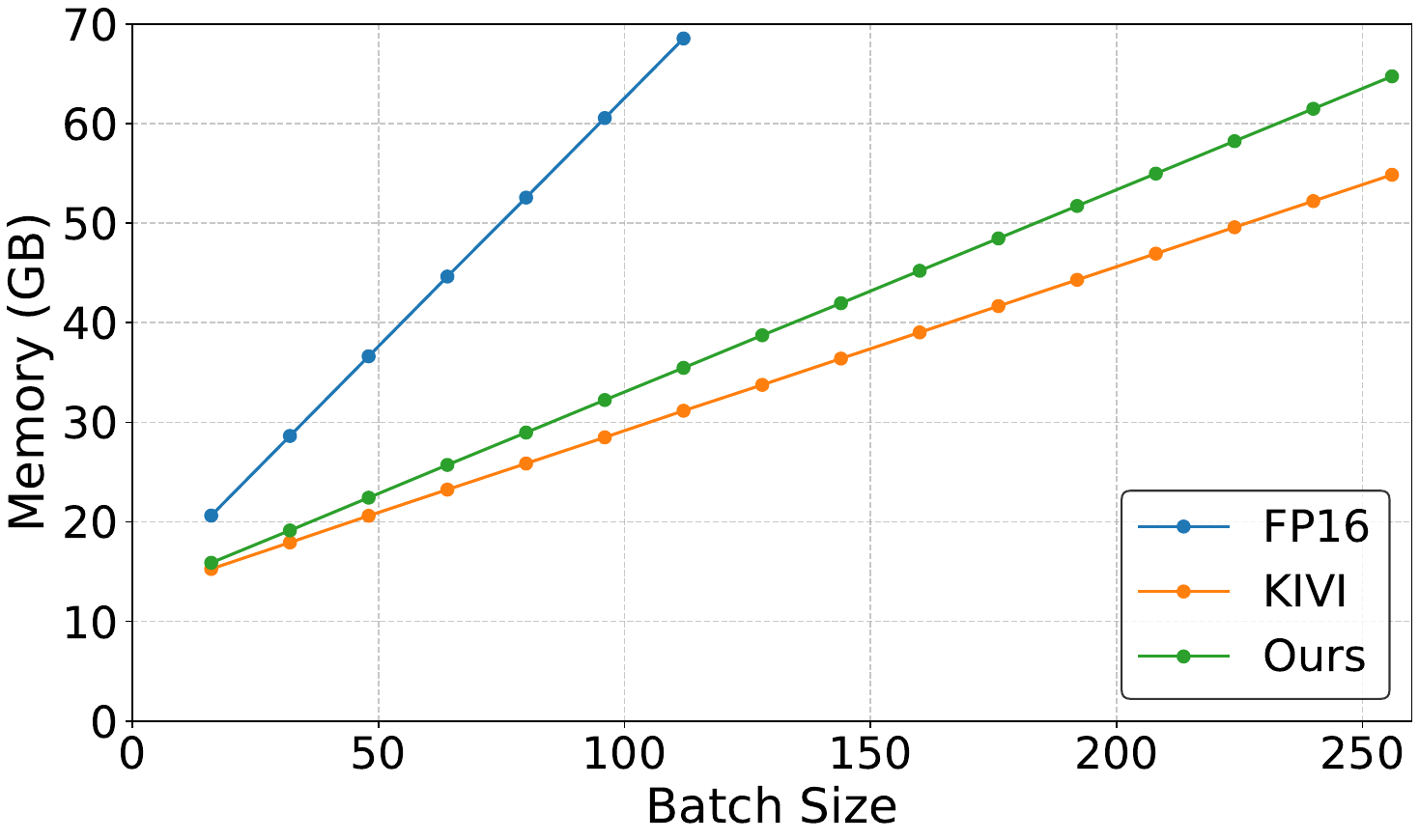}
        \caption{Memory Usage by batch size}
        \label{fig: memory by bs}
    \end{subfigure}
    \begin{subfigure}[b]{0.32\textwidth} 
        \includegraphics[width=\textwidth]{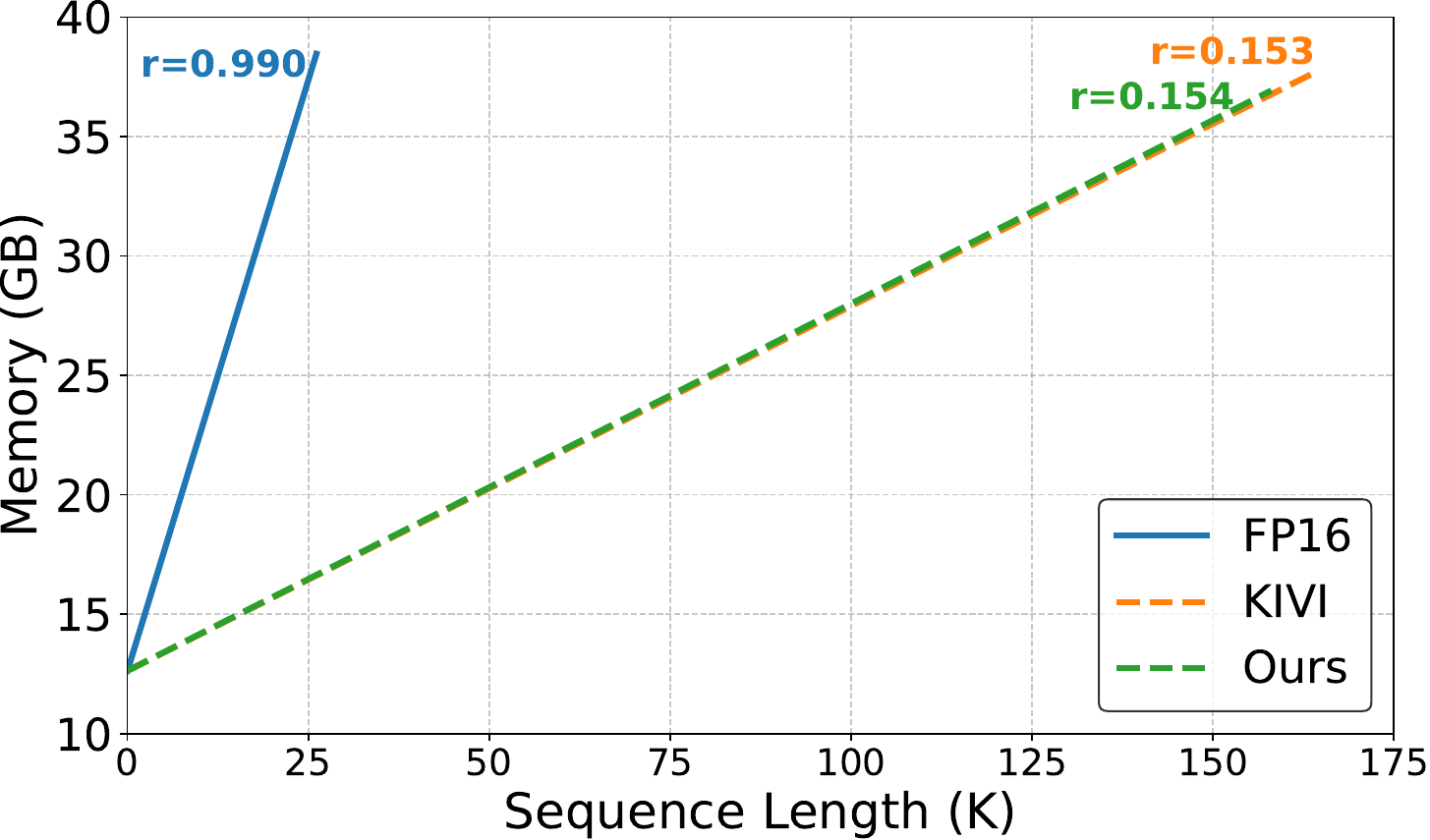}
        \caption{Memory Usage by length}
        \label{fig: memory by length}
    \end{subfigure}
    \caption{Experiments on throughput and memory: (a) Comparison of throughput (tokens/s) for different methods across different batch sizes on NVIDIA A800 80G. (b) Peak memory usage (including model weights and other components) at different batch sizes on NVIDIA A800 80G. (c) Peak memory usage (including model weights and other components) at different sequence lengths when batch size =  1 on NVIDIA A100 40G. The results shows that \textit{OTT} achieves a peak memory reduction of up to 6.4× and a throughput increase of 2.3×.}
    \label{fig: efficiency comparison}
\end{figure*}

Additionally, to validate the memory reduction and throughput improvements achieved by \textit{OTT}, we conduct three experiments: a throughput test, a memory test, and a longest sentence test.
The throughput test measures the number of tokens generated per second as the batch size varies while keeping the input and output lengths fixed. The memory test tracks memory usage as the batch size changes, also with fixed input and output lengths. The longest sentence test assesses the memory required for inference as the output length increases infinitely (until out-of-memory), with a fixed batch size of 1 and an input length of 1.
We use the LLaMA-2-7B-chat-hf model for our experiments, and set the input length to 64 and the output length to 384 for both the throughput and memory tests. Figure ~\ref{fig: efficiency comparison} illustrates the results.

Figure \ref{fig: throughput} shows that when the batch size is small, \textit{OTT} performs slightly slower than the FP16 baseline. However, as the batch size increases, \textit{OTT} demonstrates a significant speed advantage. Our method is consistently faster than KIVI at any batch size because it does not require compressing the Values at each step. Although processing outlier tokens introduces some additional computational overhead, the outlier pool is very small, and the compression frequency is low. As a result, the overhead is negligible in the overall decoding process.

From Figure \ref{fig: memory by bs}, it is evident that quantization significantly reduces memory usage compared to the FP16 baseline. \textit{OTT} requires slightly more memory than KIVI, this is because \textit{OTT} tends to retain more full-precision tokens. However, as the sequence length increases, this impact diminishes.
Figure \ref{fig: memory by length} provides a clearer view of the compression ratio (represented by the slope of each line) for the KV Cache. When the sequence length becomes sufficiently large, the effects of full-precision tokens are negligible. Notably, the compression ratio of KIVI and \textit{OTT} reaches approximately 6.4x.
We provide more time analysis in Appendix \ref{appendix:time} and \ref{additional baselines}

\subsection{Ablation studies}

\textbf{Group size and residual length.}
Group size and residual length are critical hyperparameters in \textit{OTT}. Theoretically, a larger group size allows more values to be quantized at each step, which can reduce quantization accuracy due to the increased range of $\mX_{max}-\mX_{min}$. On the other hand,  a larger group size decreases memory usage by requiring fewer quantization coefficients to be retained. Conversely, increasing the residual length requires more memory since a larger full-precision KV Cache must be retained, but it also improves accuracy. Thus, selecting an appropriate group size and residual length is critical to balancing memory usage and accuracy.
We explore the impact of group size and residual length with group sizes of \{32, 64, 128\} and residual lengths of \{0, 8, 16, 32, 64, 128\}. Table \ref{tab: Group residual} reports the results for LLaMA-3-8B-Instruct on Gsm8k 8-shot and 8-shot CoT under different configurations. When the group size is fixed, we observe a clear upward trend in accuracy as the residual length increases. However, when the residual length is fixed, the effect of group size shows no clear pattern, likely because the token distribution is relatively uniform, meaning that increasing group size has a limited impact.
Since increasing the group size can improve the compression ratio (if not consider the group tokens), we tend to choose a larger group size.
For our main experiments, we choose a group size of 128 and a residual length of 32 to balance performance and compression ratio.

\begin{table*}[t]
\
\centering
\small
\renewcommand{\arraystretch}{1.5} 
\setlength{\tabcolsep}{3.5pt} 
\resizebox{1.6\columnwidth}{!}{%
\begin{tabular}{llcc|llcc|llcc}
\toprule
$G$ & $R$ & Gsm8k(8) & Gsm8k(8-CoT) & $G$ & $R$ & Gsm8k(8) & Gsm8k(8-CoT) & $G$  & $R$ & Gsm8k(8) & Gsm8k(8-CoT) \\ \hline

32&	0	&70.05	&73.16	&64	&0	  &68.92	&72.63	&128	&0	 &70.96	&73.54 \\
32&	8	&71.95	&74.22	&64	&8	  &70.05	&73.01	&128	&8	 &72.51	&74.15 \\
32&	16	&72.78	&74.83	&64	&16	  &70.89	&73.84	&128	&16	 &72.93	&74.37 \\
32&	32	&72.78	&74.53	&64	&32	  &72.40	&75.06	&\underline{128}	&\underline{32}	 &\underline{72.55}	&\underline{75.06} \\
32&	64	&74.00	&76.88	&64	&64	  &73.69	&76.42	&128	&64	 &73.24	&75.36 \\
32&	128	&73.77	&77.33	&64	&128  &74.68	&76.42	&128	&128 &73.24	&76.65 \\
     
\bottomrule
\end{tabular}}
\caption{Results of different $G$ and $R$. 
The settings in the main experiment are indicated with underlines.} 
\label{tab: Group residual} 
\end{table*}

\textbf{The number of outlier tokens.}
We explore the effect of varying $outlier\_num$ from 0 to 6, keeping all other settings unchanged. Table \ref{sink} presents the results for LLaMA-3-8B-Instruct on Gsm8k (8-shot and 8-shot CoT). The results show that retaining even a single outlier token can significantly improve performance, but further increases in $outlier\_num$ yield diminishing returns, eventually plateauing performance.
However, the increase in $outlier\_num$ may result in more memory overhead, leading to a decrease in compression ratio.
Considering that a small $outlier\_num$ is already sufficient to significantly improve the accuracy, we set $outlier\_num = 3$ for our main experiments.
\begin{table*}[t]
    \centering
    \label{tab:total}
    \begin{subtable}[b]{0.4\textwidth}
        \centering
        \small
        \renewcommand{\arraystretch}{1.2} 
        \setlength{\tabcolsep}{3.0pt} 
        \begin{tabular}{ccc}
        \toprule
        \multicolumn{1}{c}{$outlier\_num$} & \multicolumn{1}{c}{Gsm8k(8)} & \multicolumn{1}{c}{Gsm8k(8-CoT)} \\ \hline
        0	&62.09	&68.31 \\
        1	&71.80	&75.74 \\
        2	&71.57	&75.06 \\
        \underline{3}	&\underline{72.55}	&\underline{75.06} \\
        4	&72.25	&75.97 \\
        5	&72.18	&75.74 \\
        6	&72.18	&75.89 \\
        \bottomrule
        \end{tabular}
    \end{subtable}
    \begin{subtable}[b]{0.4\textwidth}
        \centering
        \small
        \renewcommand{\arraystretch}{1.2} 
        \setlength{\tabcolsep}{3.0pt} 
        \begin{tabular}{ccc}
        \toprule
        \multicolumn{1}{c}{Layers} & \multicolumn{1}{c}{Gsm8k(8)} & \multicolumn{1}{c}{Gsm8k(8-CoT)} \\ \hline
        None	  & 72.48	& 75.59 \\
        0	      & 72.78	& 75.44 \\
        \underline{0,1}	      & \underline{72.55}	& \underline{75.06} \\
        $0\sim 2$ & 71.49	& 74.68 \\
        $0\sim 3$ & 71.80	& 73.64 \\
        $0\sim 4$ & 70.96	& 74.53 \\
        $0\sim 5$ & 69.60	& 74.00 \\
        \bottomrule
        \end{tabular}
    \end{subtable}
        \caption{Ablation study of $outlier\_num$. The settings in the main experiment are indicated with underlines. Left: results on Gsm8k with different $outlier\_num$. Right: results on Gsm8k with $outlier\_num=0$ in shallow layers.}
    \label{sink}
\end{table*}

\textbf{Outlier tokens in shallow layers.}
\label{sinknum}
We observe that there are no outlier tokens in the shallow layers (see Figure \ref{appendix_layer} and Figure \ref{fig:appendix_channel} in Appendix, the Keys in the shallow layers does not exhibit the characteristics discussed in Section \ref{outlier}), suggesting that $outlier\_num$ should be set to 0 in these layers. To explore this further, we set $outlier\_num$ to 0 in consecutive shallow layers and evaluate the performance on Gsm8k (8-shot and 8-shot CoT) using LLaMA-3-8B-Instruct. For example, ``$0\sim 2$'' means that $outlier\_num$ is set to 0 for the first three layers of the model. Table \ref{sink} shows that the impact is minimal in the shallowest layers but becomes more significant as we move deeper into the model. Based on these results, we set $outlier\_num = 0$ for the first two layers in all models for our main experiments.



\section{Related work}
\label{Section: 5}

\textbf{Efficient Inference of LLMs.} \ 
Large Language Models often have enormous parameters, leading to significant computational costs on inference. To address this, some researchers have employed parameter pruning techniques to eliminate redundant or less important parameters, thereby compressing LLMs~\citep{LLM_Pruner,ShearedLLaMA,SparseGPT}. Other studies have focused on quantizing model weights, reducing their size and the number of arithmetic operations required for inference. For example, GPTQ~\citep{GPTQ} uses second-order information to quantize models to 3 or 4-bit precision while maintaining accuracy. AWQ~\citep{AWQ} preserves critical weights based on the activation distribution, quantizing the remaining weights to lower bit precision. These methods can be combined with KV Cache compression to achieve a better performance.

\textbf{KV Cache Compression.} \ 
KV Cache compression can significantly reduce the size of KV Cache with minimal accuracy loss.
\citet{KIVI} find that some outlier channels in the Keys have very large magnitudes, resulting in a significant loss. \citet{KV_QUANT} find that quantizing the Key cache before applying rotary positional embeddings reduces the negative impact of quantization. \citet{Streamingllm} propose StreamingLLM, which retains the initial and final tokens of the input. 
Similarly, \citet{sun2024massive} find a "massive activations" pattern in LLMs, where a few activations have much higher values than others. These values stay stable across inputs and act as critical bias terms in the model. \citet{H20} find that only a minority of tokens influence the output.
\section{Conclusion}
In this paper, we start from the assumptions of KIVI and further explore the distribution of the Keys in the outlier channels.
We observe that a few outlier tokens deviate from the assumptions of KIVI.
Quantizing these tokens has detrimental effects, as it increases the quantization errors of other tokens.
Building on these observations, we propose KV Cache Quantization with Outlier Tokens Tracing (\textit{OTT}), which leverages the magnitude of the Keys to dynamically trace these tokens during decoding, excluding them from the quantization process while retaining their full-precision representations.
Extensive experiments show that our method achieves significant improvements in accuracy, along with substantial reductions in memory usage and increases in throughput.

\section*{Limitations}
Although \textit{OTT} has achieved excellent results, there are still some limitations:
\begin{itemize}
    \item Due to the presence of the group and nearest tokens, we cannot ensure that all tokens are quantized at every moment. When the sequence length is very short and the batch size is very large, the compression ratio of \textit{OTT} is reduced. In extreme cases, when the sequence length is shorter than the group size, \textit{OTT} does not perform any compression.
    \item \textit{OTT} occasionally still incurs a little loss on specific datasets under 2-bit quantization. This may be related to the difficulty of the datasets and the required generation length. When the generation length is very long, \textit{OTT} may face an unacceptable risk of loss due to error accumulation.
\end{itemize}

\section*{Acknowledgements}
We want to thank all the anonymous reviewers for theirvaluable comments. This work was supported by the National Science Foundation of China (NSFC No.62206194)the Natural Science Foundation of Jiangsu Province, China(Grant No, BK20220488), the Young Elite Scientists Spon.sorship Program by CAST (2023QNRC001), and the Pri-ority Academic Program Development of Jiangsu lligherEducation Institutions.

\bibliography{custom}

\appendix
\onecolumn
\clearpage
\section{Additional Figures and Tables}

\begin{table*}[!ht]
    \centering
    \begin{tabular}{|p{14cm}|}  
        \hline
        Input: \\
        \quad\quad[INST]Which methods did Socrates employ to challenge the prevailing thoughts of his time?\\
        \quad\quad [/INST] \\
        \hline
        Output: \\
        \quad\quad[INST]Which methods did Socrates employ to challenge the prevailing thoughts of his time?\\
        \quad\quad[/INST] Socrates employed several methods to challenge the prevailing thoughts of his time, including:\\
        
        \quad\quad1. Questioning: Socrates used questioning as a primary method to challenge the beliefs of others. He would ask probing questions to expose the contradictions and inconsistencies in their arguments, and to lead them to a deeper understanding of their own beliefs.\\
        \quad\quad2. Dialogue: Socrates believed in the power of dialogue to promote critical thinking and understanding. He would engage in conversations with others, using his questions and their responses to challenge their beliefs and to encourage them to think more deeply about their ideas.\\
        \quad\quad3. Irony: Socrates often used irony to challenge the beliefs of others. By using irony, he could expose the flaws in their arguments and lead them to see things from a different perspective.\\
        \quad\quad4. Paradox: Socrates often used paradoxical statements to challenge the prevailing beliefs of his time. By presenting seemingly contradictory statements, he could encourage his interlocutors to think more critically about their beliefs and to question their assumptions.\\
        \quad\quad5. Maieutics: Socrates believed in the power of midwifery, or maieutics, to bring forth new ideas and insights. He would use his questions and dialogue to help others discover new truths and to challenge their existing beliefs.\\
        \quad\quad6. Analogy: Socrates often used analogies to help others understand complex ideas and to challenge their beliefs. By using comparisons and contrasts, he could help his interlocutors to see things from a different perspective and to develop a deeper understanding of the issues at hand.\\
        \quad\quad7. Refutation: Socrates was skilled at refuting the arguments of others by exposing their flaws and inconsistencies. He would use his questions and dialogue to challenge the beliefs of others and to lead them to a deeper understanding of the issues at hand.\\
        \quad\quad8. Affirmation: Socrates also used affirmation to challenge the prevailing thoughts of his time. By affirming the beliefs of others and building on them, he could encourage them to think more deeply about their ideas and to develop a more nuanced understanding of the issues at hand.\\
        \quad\quad9. Silence: Socrates believed that sometimes the best way to challenge the prevailing thoughts of his time was through silence. By remaining silent in the face of foolish or false arguments, he could encourage others to think more critically about their beliefs and to question their assumptions.\\
        
        \quad\quad Overall, Socrates employed a wide range of methods to challenge the prevailing thoughts of his time, including questioning, dialogue, irony, paradox, maieutics, analogy, refutation, affirmation, and silence. Through these methods, he was able to encourage critical thinking and to promote a deeper understanding of the issues at hand.\\
        \hline
    \end{tabular}
    \caption{Example generated by LLaMA-2-7B-chat-hf.}
    \label{tab:example}
\end{table*}

\clearpage
\begin{figure*}[!htbp]
    \centering
    \includegraphics[width=1\linewidth]{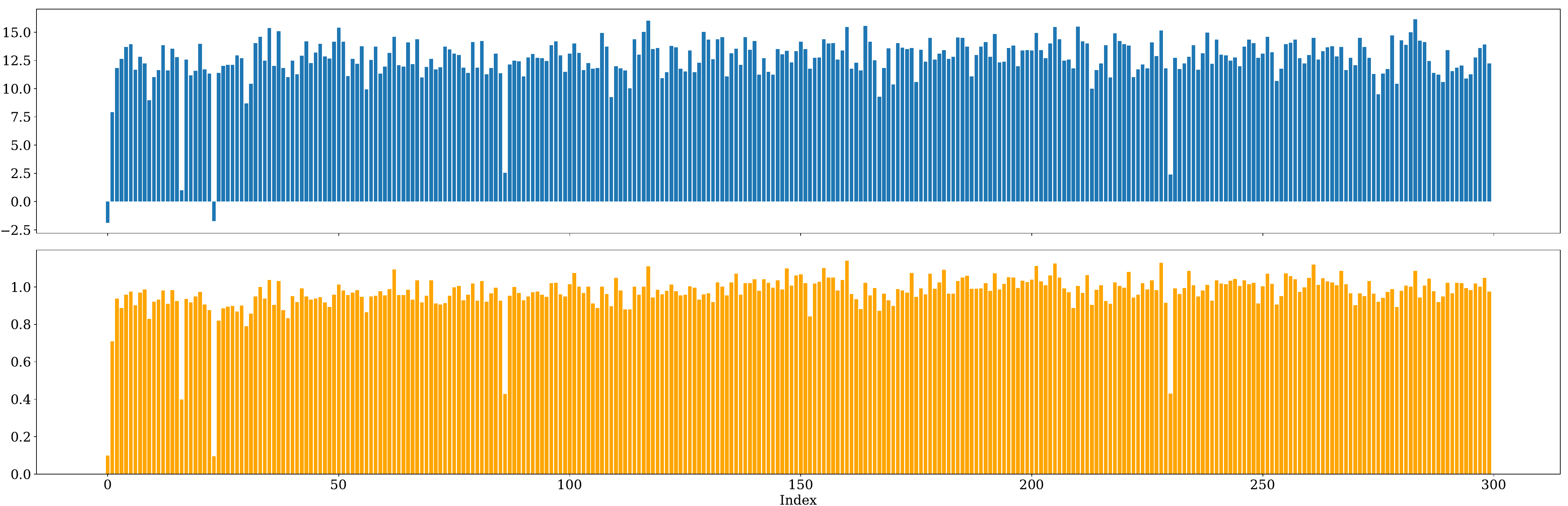}
    \caption{The Keys in an outlier channel (up) and the magnitude of the Keys overall (down).}
    \label{fig:compare}
\end{figure*}

\begin{figure*}[!htbp]
  \centering
  \begin{subfigure}[b]{0.24\textwidth}
    \caption*{Layer 0 key cache}
    \includegraphics[width=\textwidth]{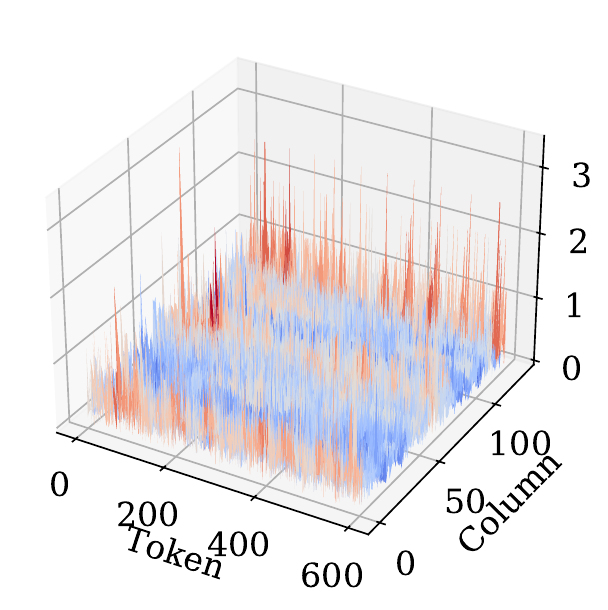}
  \end{subfigure}
  \begin{subfigure}[b]{0.24\textwidth}
    \caption*{Layer 0 value cache}
    \includegraphics[width=\textwidth]{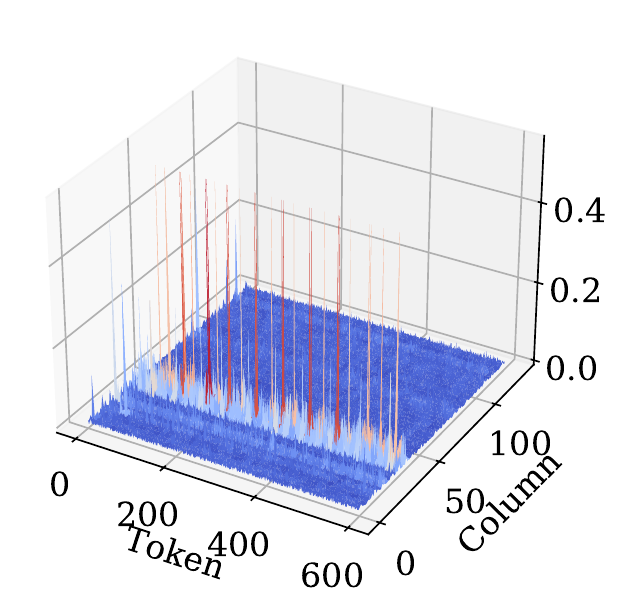}
  \end{subfigure}
  \begin{subfigure}[b]{0.24\textwidth}
    \caption*{Layer 10 key cache}
    \includegraphics[width=\textwidth]{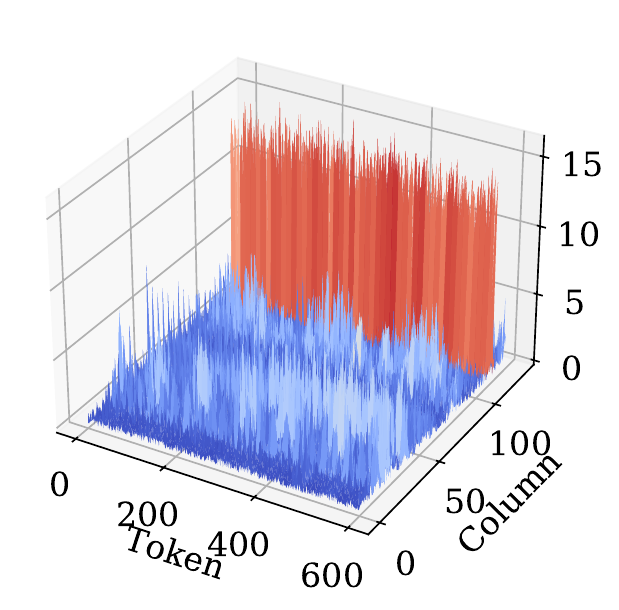}
  \end{subfigure}
  \begin{subfigure}[b]{0.24\textwidth}
    \caption*{Layer 10 value cache}
    \includegraphics[width=\textwidth]{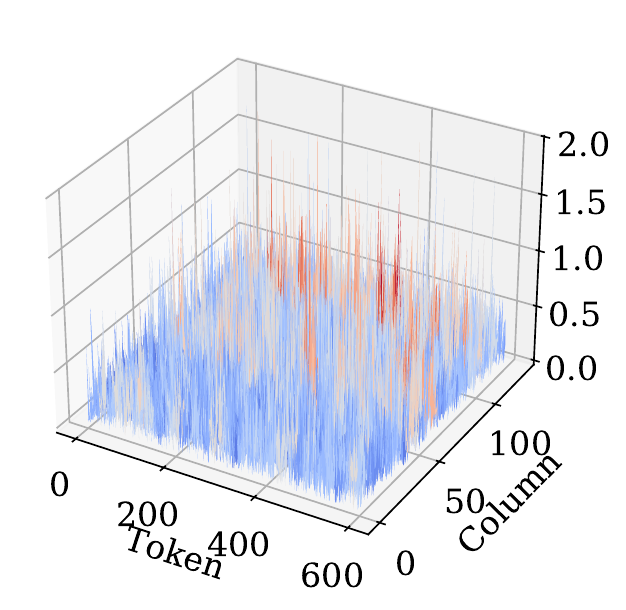}
  \end{subfigure}
  
  \begin{subfigure}[b]{0.24\textwidth}
    \caption*{Layer 20 key cache}
    \includegraphics[width=\textwidth]{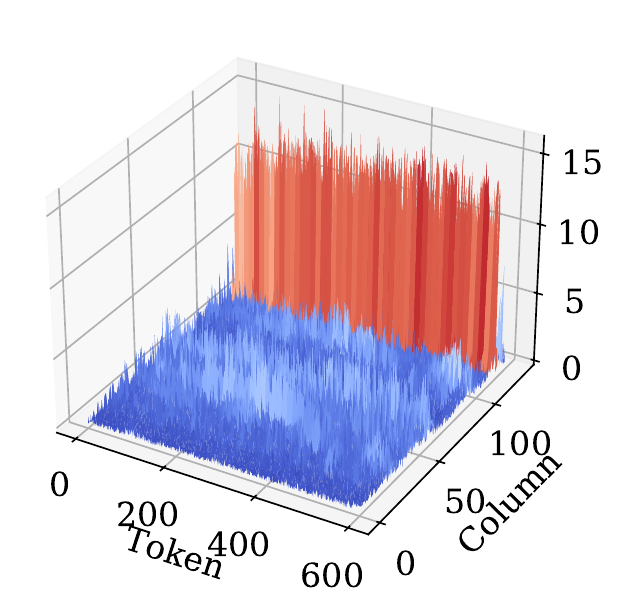}
  \end{subfigure}
  \begin{subfigure}[b]{0.24\textwidth}
    \caption*{Layer 20 value cache}
    \includegraphics[width=\textwidth]{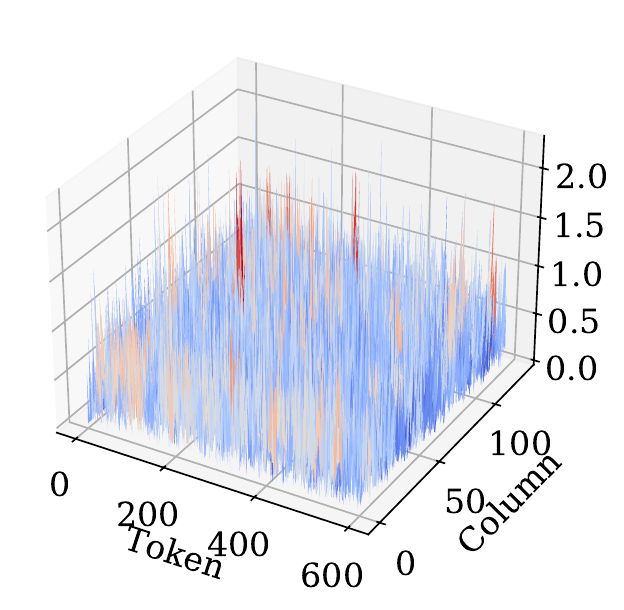}
  \end{subfigure}
  \begin{subfigure}[b]{0.24\textwidth}
    \caption*{Layer 31 key cache}
    \includegraphics[width=\textwidth]{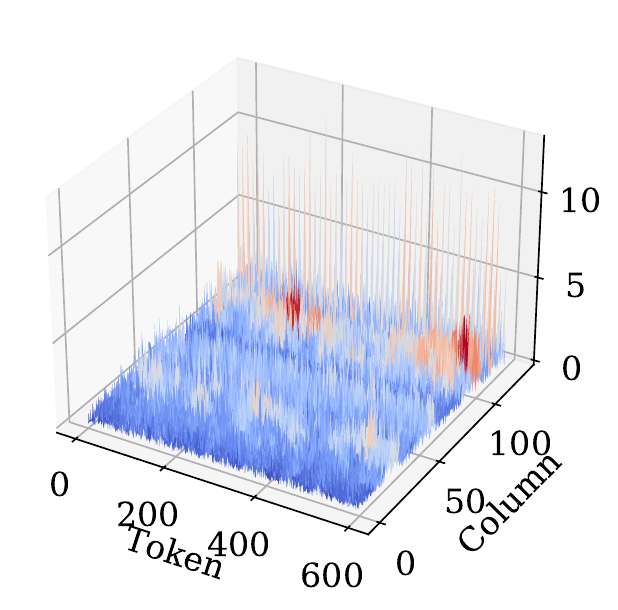}
  \end{subfigure}
  \begin{subfigure}[b]{0.24\textwidth}
    \caption*{Layer 31 value cache}
    \includegraphics[width=\textwidth]{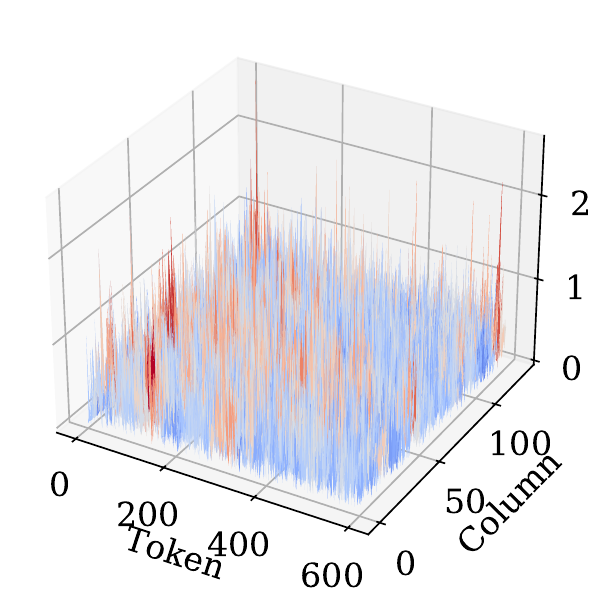}
  \end{subfigure}
  \caption{Magnitude of the keys and Values for Llama-2-7B-chat-hf in head 17.}
  \label{appendix_layer}
\end{figure*}

\begin{figure*}[!htbp]
    \centering
    \includegraphics[width=1\linewidth]{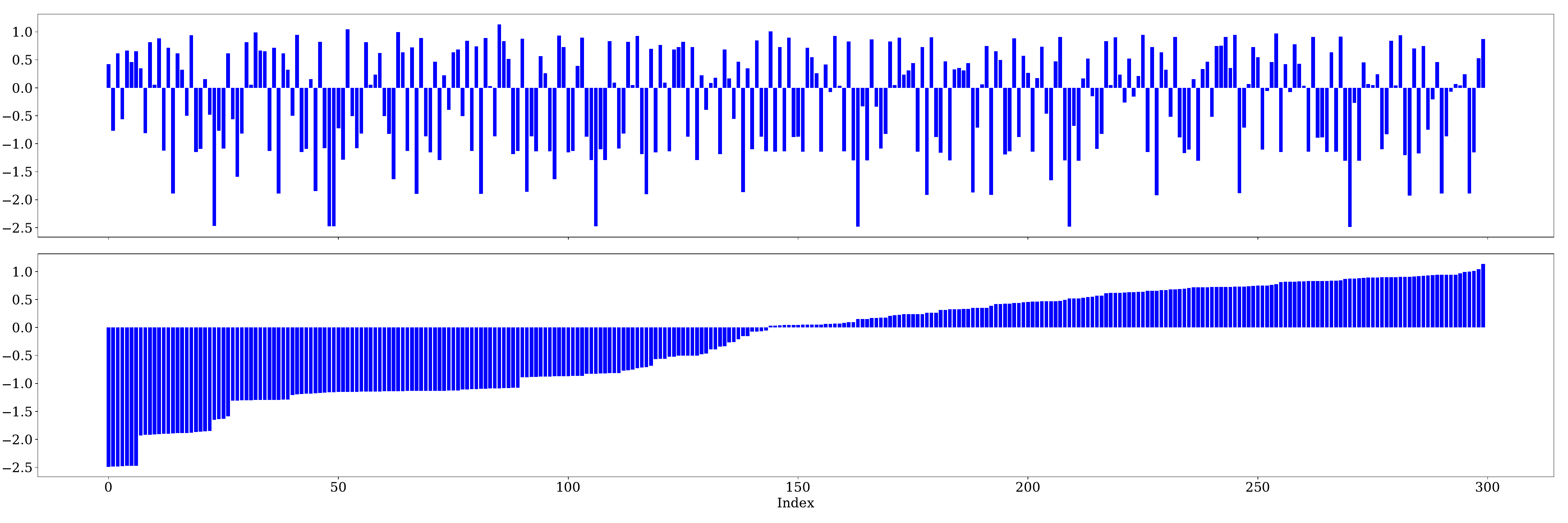}
    \caption{The Keys in an outlier channel (up) and the sorted Keys in an outlier channel (down).}
    \label{fig:appendix_channel}
\end{figure*}

\clearpage
\twocolumn
\section{Additional Experiment Results}
\label{additional results}
\subsection{Experiments on LLaMA-2-70b-chat-hf}
To validate the performance on larger models, we conduct additional experiments on LLaMA-2-70b-chat-hf. 
The experimental setup is completely consistent with the main experiment. The result in Table \ref{tab:70b} shows that \textit{OTT} can still achieve higher accuracy advantages on larger models based on KIVI.
\begin{table}[!htbp]
\centering
\small
\renewcommand{\arraystretch}{1.6} 
\setlength{\tabcolsep}{4.2pt} 
\resizebox{1\columnwidth}{!}{%

\begin{tabular}{l|ccccccc}
\toprule
70b-chat-hf & Gsm8k(8) & Gsm8k(8-cot) & Gsm8k(0-cot) & BBH(3) & HE(p@1) & Avg   \\ \hline
FP16               & 56.03    & 55.04        & 48.98        & 47.09  & 16.46                      & 44.72 \\
KIVI              & 51.63    & 50.49        & 46.40        & 46.08  & 14.02                            & 41.72 \\
Ours               & \textbf{52.92}    &  \textbf{52.54}        &  \textbf{49.05}        &  \textbf{46.48}  &  \textbf{15.85}                   &  \textbf{43.37} \\
\bottomrule
\end{tabular}}
\caption{Experiments on LLaMA-2-70b-chat-hf}
\label{tab:70b} 

\end{table}

\subsection{Comparison with token eviction methods}
We add some comparisons with the token eviction methods.
The previous token eviction methods are mostly evaluated on LongBench, so we also conduct experiments on LongBench using LLaMA-2-7b-chat-hf.
The input length of LongBench is relatively long, while the output length is relatively short, which may be more conducive to the performance of the token eviction methods.
The baselines include StreamingLLM \citep{Streamingllm}, H2O \citep{H20}, and SnapKV \citep{snapkv}. In order to maintain the simplicity and consistency of the settings for comparison, we only perform token eviction in the prefill stage, and retain all KV caches in the decode stage.
In addition, we make some adjustments to H2O based on SnapKV's strategy, selecting only the queries in the sliding window for attention score selection (which was later verified to be superior to H2O's strategy). In order to maintain the overall compression ratio consistent with \textit{OTT}, we choose to evict 84\% of the tokens in the prefill stage, which have the closest compression ratio to \textit{OTT}. For H2O, the number of recent tokens and heavy hitters is the same. For StreamingLLM, we do not adjust its position id during decoding phase.
So, the process of token eviction is as follows:
\begin{itemize}
    \item In the prefill stage, use queries in the sliding window to calculate the attention score with other tokens, and perform token eviction according to the strategies of StreamingLLM, H2O, and SnapKV respectively.
    \item During the decode phase, attention calculation is performed directly without token eviction.
\end{itemize}
The results are shown in Table \ref{tab:other baselines in longbench}
Among these methods, SnapKV achieves the best results. But even under more favorable settings, the result is still slightly lower than \textit{OTT}.

\subsection{Comparison with ZipCache}
We compare \textit{OTT} with ZipCache \citep{he2024zipcache}, and in order to maintain consistent compression rates, we set 20\% of the tokens to 4-bit quantization and 80\% to 2-bit quantization. The other hyper-parameters in ZipCache are the same.
We conduct our experiments on GSM8K, BBH and HumanEval with LLaMA-2-7b-chat-hf.
The results in Table \ref{tab:zipkvquant} show that ZipCache is weaker than KIVI and \textit{OTT}.

\section{Additional Time Analysis}
\label{appendix:time}
\begin{figure}[htbp]
    \centering
    \includegraphics[width=1\linewidth]{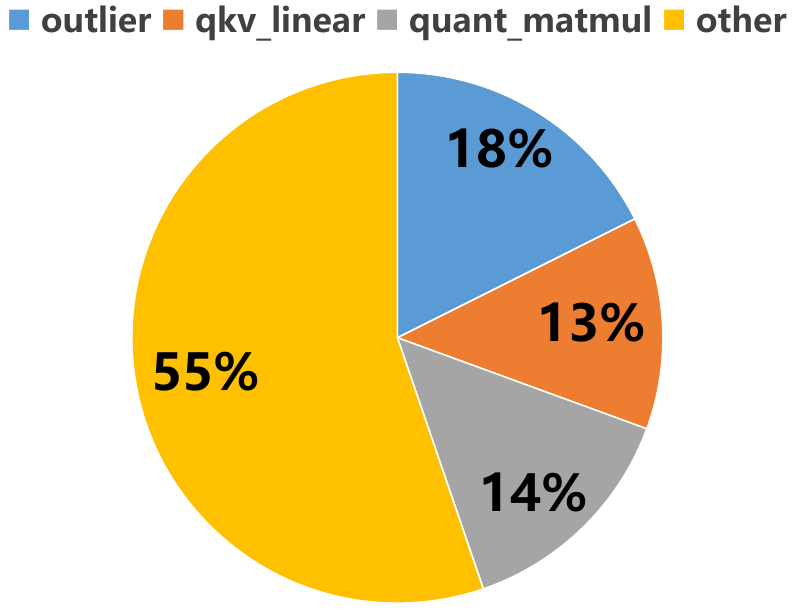}
    \caption{The time proportion in the attention block.}
    \label{fig:time}
\end{figure}

We provide a more detailed analysis of the time cost.
The computational overhead comes from two aspects.
In the compression stage, we calculate the magnitude of each token's key, perform comparison, select the index, and quantify it. The cost of the outlier operation is relatively high compared to quantization, but the compression is only performed every $G$ steps, so this time cost can be almost negligible compared to the whole decoding process.
In the attention calculation stage, we need to calculate the qkv in the outlier pool and cover the attention score according to the outlier token index, which has a certain cost.
We plot the detailed time consumption in the attention block in Figure \ref{fig:time}, and the outlier operation accounts for about 18\%.
Considering the pre-processing, post-processing and FFN calculation in the entire forward step, the time proportion of outlier operations is very small.

\begin{table*}[!htbp]
\centering
\small
\renewcommand{\arraystretch}{1.4} 
\setlength{\tabcolsep}{4.2pt} 

\begin{tabular}{lccccccccc}
\toprule
LLaMA-2-7b-chat & Qasper         & GovReport      & MultiNews      & TREC           & TriviaQA       & SamSum        & LCC           & Repobench-P    & Avg            \\ \hline
FP16           & 20.04          & 25.08          & 23.02          & 59.67          & 85.39          & 39.28         & 59.59         & 48.04          & 45.01          \\
KIVI           & \textbf{20.43} & 19.97          & 19.82          & 59.67          & \textbf{85.16} & 37.7          & 58.73         & 47.24          & 43.59          \\
SnapKV         & 18.96          & 18.73          & 19.64          & 59             & 84.84          & 38.22         & \textbf{60.5} & \textbf{50.08} & 43.75          \\
H2O            & 17.51          & 18.85          & 19.88          & 50             & 84.22          & 38.09         & 58.23         & 49.66          & 42.05          \\
Streaming      & 15.31          & 19.39          & 18.99          & 51             & 83.11          & 36.8          & 57.57         & 47.33          & 41.19          \\
Ours           & 19.95          & \textbf{21.56} & \textbf{20.81} & \textbf{59.67} & 85             & \textbf{39.1} & 59.44         & 48.51          & \textbf{44.26} \\ \bottomrule
\end{tabular}
\caption{Experiments on three additional eviction-based methods on LLaMA-2-7b-chat-hf.}
\label{tab:other baselines in longbench} 
\end{table*}

\begin{table*}[!htbp]
\centering
\small
\renewcommand{\arraystretch}{1.4} 
\setlength{\tabcolsep}{4.2pt} 

\begin{tabular}{lccc|ccc|cc|l}
\toprule
LLaMA-2-7b-chat  & Gsm8k(8)             & 8-cot          & 0-cot          & BBH(3)         & 3-cot                       & 0-cot          & HE(p@1)        & p@10                                  & \multicolumn{1}{c}{Avg} \\ \hline
FP16     & 21.99                & 21.3           & 24.11          & 33.34          & 40.21                       & 35.00          & 12.19          & 17.07                                 & 45.01                   \\
KIVI     & 16.3                 & 17.51          & 21.61          & 32.48          & 34                          & 33.30          & 9.75           & {\color[HTML]{262626} 12.19}          & 43.59                   \\
Ours     & \textbf{19.86}       & \textbf{19.33} & \textbf{22.52} & \textbf{33.33} & \textbf{34.43}              & \textbf{33.74} & \textbf{11.58} & {\color[HTML]{262626} \textbf{14.63}} & \textbf{43.75}          \\
ZipCache & 15.92                & 17.74          & 20.02          & 32.85          & {\color[HTML]{262626} 33.9} & 32.35          & 9.45           & 15.24                                 & 42.05                   \\
\bottomrule
\end{tabular}
\caption{Experiments on ZipCache.}
\label{tab:zipkvquant} 
\end{table*}

\section{Proof of Low-Magnitude Keys Disrupting Attention Weights}
We formalize the claim that \textit{low-magnitude keys in outlier channels disrupt attention weights} through two steps:

\subsection{Quantization Error}
\begin{itemize}
    \item \textbf{Definitions}:
    \begin{itemize}
        \item Let $K_c \in \mathbb{R}^{n}$ be an outlier channel containing $n$ Key values where $\exists$ a subset $S \subset \{1,\ldots,n\}$ with $|S|=m \ll n$ such that:
        $$
        \begin{cases}
        K_{c,i} \in [\mu-\sigma, \mu+\sigma],  \forall i \notin S \\ \text{ (uniform distribution)} \\
        K_{c,j} \in [\epsilon, \delta],  \forall j \in S \\ \text{ where } 0 < \epsilon \ll \mu-\sigma 
        \end{cases}
        $$
    \end{itemize}

    \item \textbf{Quantization Parameters}:
    \begin{itemize}
        \item Full range: $X_{\text{max}} = \max(K_c),\ X_{\text{min}} = \min(K_c)$
        \item Quantization step: $q = \frac{X_{\text{max}}-X_{\text{min}}}{2^b-1}$
    \end{itemize}

    \item \textbf{Key Observation}:
    The presence of low-magnitude outliers forces:
    $$
    X_{\text{min}} \leq \epsilon \ll \mu-\sigma \quad \text{and} \quad X_{\text{max}} \geq \mu+\sigma
    $$
    $$
    \implies q = \frac{(\mu+\sigma)-\epsilon}{2^b-1} \gg \frac{2\sigma}{2^b-1}
    $$
    
    \item \textbf{Result}: Low-magnitude outliers inflate quantization step size, leading to larger approximation errors for \textit{all tokens} in the channel.
\end{itemize}

\subsection{Error Propagation to Attention Weights}
\begin{itemize}
    \item \textbf{Attention Score Calculation}:
    For query vector $Q \in \mathbb{R}^d$ and quantized Key matrix $K'$:
    $$
    A_i = \frac{QK_i'}{\sqrt{d}} \quad \text{and} \quad A_i^{\text{quant}} = A_i + \underbrace{\frac{Q(K_i'-K_i)}{\sqrt{d}}}_{\Delta A_i}
    $$

    \item \textbf{Error Analysis}:
    \begin{itemize}
        \item For outlier channel $c$:
        $$
        \mathbb{E}[|K_{c,i}'-K_{c,i}|] \propto q 
        $$
        $$
        \Delta A_i \sim \sum_{c=1}^d Q_c(K_{c,i}'-K_{c,i})
        $$
    \end{itemize}

    \item \textbf{Key Observation}: In outlier channels where $|Q_c|$ is typically large (by definition of being ``outlier channels''), the quantization error gets amplified by:
    $$
    \Delta A_i \propto Q_c(K_{c,i}'-K_{c,i}) \approx Q_c \cdot q
    $$
    
    \item \textbf{Result}: The error in quantization steps propagates to attention weights.
\end{itemize}

\section{Mathematical Formulation of OTT}
\label{math}

\paragraph{Outlier Token Identification}

For a token \( t_i \) with Key vector \( \boldsymbol{K}_i \), its outlier score \( S_i \) is computed as the magnitude of its Keys, typically measured via the \( L_1 \)-norm:  
\[
S_i = \|\boldsymbol{K}_i\|
\]
Tokens with smaller \( S_i \) are identified as outliers since their Keys deviate significantly from the uniform distribution in outlier channels.

\paragraph{Competition Mechanism}

At each quantization step (every \( G \) tokens), tokens in the current group \( \mathcal{T} = \{ t_1, t_2, \dots, t_G \} \) compete with the existing outlier pool \( \mathcal{O} \) (capacity \( N \)) for inclusion. The process involves:

\begin{enumerate}
    \item \textbf{Score Calculation:} Compute \( S_i \) for all tokens in \( \mathcal{T} \) and \( \mathcal{O} \).
    \item \textbf{Token Ranking:} Combine \( \mathcal{T} \) and \( \mathcal{O} \), then sort all tokens by \( S_i \) in ascending order:
    \[
    \text{Sorted List} = \text{argsort}(S_i), \quad \forall t_i \in \mathcal{T} \cup \mathcal{O}
    \]
    \item \textbf{Outlier Pool Update:} Select the top-\( N \) tokens with the smallest \( S_i \) to form the new outlier pool:
    \[
    \mathcal{O}_{\text{new}} = \left\{ t_j \mid j \in \text{top-}N \text{ indices of Sorted List} \right\}
    \]
    \item \textbf{Replacement Handling:} Tokens evicted from \( \mathcal{O} \) (if \( |\mathcal{T} \cup \mathcal{O}| > N \)) are stored in an auxiliary pool or discarded. The retained outlier tokens are excluded from quantization and stored in full precision.
\end{enumerate}

\paragraph{Mathematical Formulation}

Let \( \mathcal{O}^{(t)} \) denote the outlier pool at step \( t \), and \( \mathcal{T}^{(t)} \) the current token group. The update rule is:  
\[
\mathcal{O}^{(t+1)} = \underset{\substack{\text{top-}N \text{ tokens by } S_i}}{\arg\min} \left( \mathcal{O}^{(t)} \cup \mathcal{T}^{(t)} \right)
\]
Outlier tokens are excluded from quantization, while non-outliers are quantized using channel-wise (Keys) and token-wise (Values) methods as in KIVI.

\section{Statistical Analysis of Preliminary Results}
\label{statistical}
We conduct additional analysis experiments on LongBench using LLaMA-2-7b-chat-hf. We record the keys of Layer 10, Head 16 of the first 1024 tokens during the generation process and analyze the distribution of the outlier channel (i.e., the channel with the largest magnitude) among these keys. Specifically, for each example, we identify the outlier channel of the key in Layer 10, Head 16, then divide the entire range of channel values into ten equal parts and record which range each token's value fall into. Finally, we average the results across the entire dataset. The results are shown in Table \ref{tab:statistical}, which confirms our previous hypothesis.

\section{Additional Benchmarks and Baselines}
\label{additional baselines}
To validate the effectiveness of OTT, we also add other baselines and benchmarks. Among them, the benchmarks include Needle-in-a-Haystack and Ruler, while the baselines include ZipCache and Gear.

\textbf{RULER} ~\citep{hsieh2024ruler}: This benchmark evaluates models' ability to handle complex reasoning tasks. It involves tasks that require understanding and linking various pieces of information, making it essential for assessing skills in multi-step reasoning and logical analysis.

\textbf{Needle-in-a-Haystack} ~\citep{liu2024lost}: This benchmark focuses on testing if models can find important details in long texts. It checks how well models can spot useful information in a lot of text, which is key for tasks like finding facts or answering questions by pulling out parts of the text.
 
\textbf{GEAR} \citep{GEAR} compensates for compression-induced errors by combining low-rank and sparse matrices, achieving near-lossless results in 2-bit quantization integrated with KIVI \citep{KIVI}. 

\textbf{ZipCache} \citep{he2024zipcache} achieves accurate KV cache compression by introducing a channel-separable tokenwise quantization scheme, an improved salient token identification metric based on normalized attention scores, and an efficient approximation method for fast attention implementations.

We adjust the hyper-parameters of various methods to thoroughly observe their performance.
For KIVI and OTT, we vary the group size $G$ and the residual length $R$.
For ZipCache, we vary the k unimportant ratio $k$ and v unimportant ratio $v$.
The unimportant tokens are stored in 2-bit and the important ones are stored in 4-bit.
For Gear, we set the low rank $r$ to 2, group size $G$ to 128, streaming gap to 100, outlier ratio to 0.01. Note that Gear will use much more memory and is much slower than KIVI and OTT because it add additional low-rank and outlier operations based on KIVI.
\subsection{Results on Needle-in-a-Haystack}
Figure \ref{fig:needle_res1} and \ref{fig:needle_res2} shows the results of different methods and models on Needle-in-a-Haystack.
The results on LLaMA-3-8B-Instruct shows that all methods can perform well under this setting.
The results on LLaMA-2-7B-chat-hf show that Gear performs the best across all methods, while it sacrifices memory and throughput. OTT performs better than KIVI (increasing the accuracy from 93.1\% to 99.2\%). ZipCache performs the worst.

\subsection{Results on Ruler}
We validate the effectiveness of various methods on Ruler. The results are shown in Table \ref{table:ruler}.
Similar to previous findings, Gear still achieves the best accuracy among all methods, thanks to its higher computational cost and memory usage.
OTT achieves better accuracy than KIVI when $G=32$ and $G=128$, demonstrating the effectiveness of our method.
ZipCache also achieve the worst results, with significant losses on both models.

\subsection{Full Results on LongBench}

We test the performance of each method on LongBench under more settings and complete all the results for LongBench.
The results are shown in Table \ref{table:longbench}.
OTT achieves almost no loss on LongBench, performing nearly as well as Gear, and clearly outperforming ZipCache and KIVI.

\subsection{Results on Helmet}
We test the performance of each method on Helmet \citep{helmet} benchmark using LLaMA-3-8B-Instruct. We set the max length to 8192. The results are shown in Table \ref{tab:helmet}. The results show that our method performs better than the baselines under fair comparison.

\subsection{Results on LongBench-v2}
We test the performance of each method on LongBench-v2 \citep{longbenchv2} using LLaMA-3-8B-Instruct. We set the max length to 8192. The results are shown in Table \ref{tab:longbench2}. The results show that our method performs better than the baselines under fair comparison.

\subsection{Results on Longer Models}
We test the performance of each method on Ruler with Llama-3-8B-ProLong-512k-Instruct and longer context lengths (64k, 128k), the results are shown in Table \ref{table:longer_ruler}.
We can conclude from the tables that our method can perform well on extreme-long scenarios.

\subsection{Throughput and Memory Analysis}
\begin{figure*}[htbp]
  \centering
  \begin{subfigure}[b]{0.48\textwidth}
    \includegraphics[width=\textwidth]{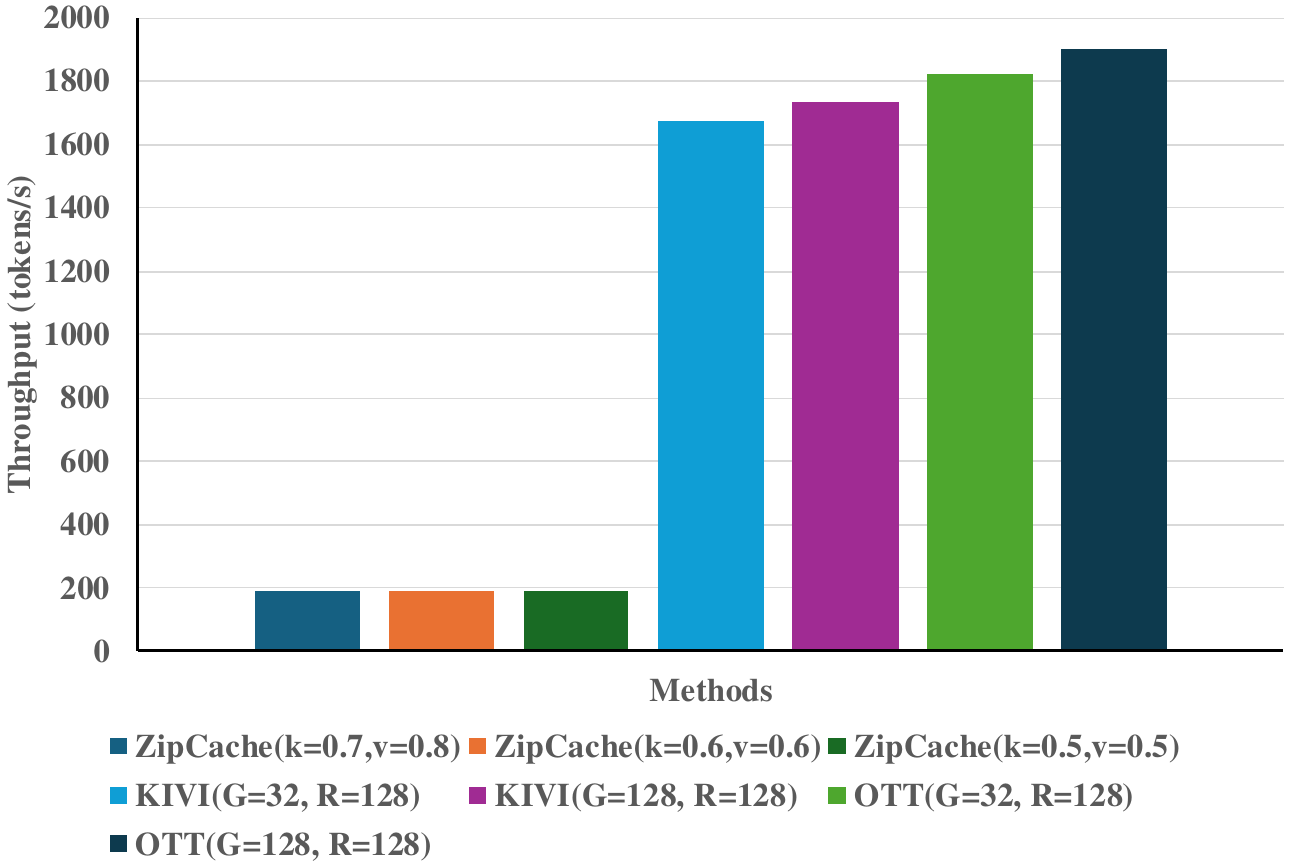}
  \end{subfigure}
  \begin{subfigure}[b]{0.48\textwidth}
    \includegraphics[width=\textwidth]{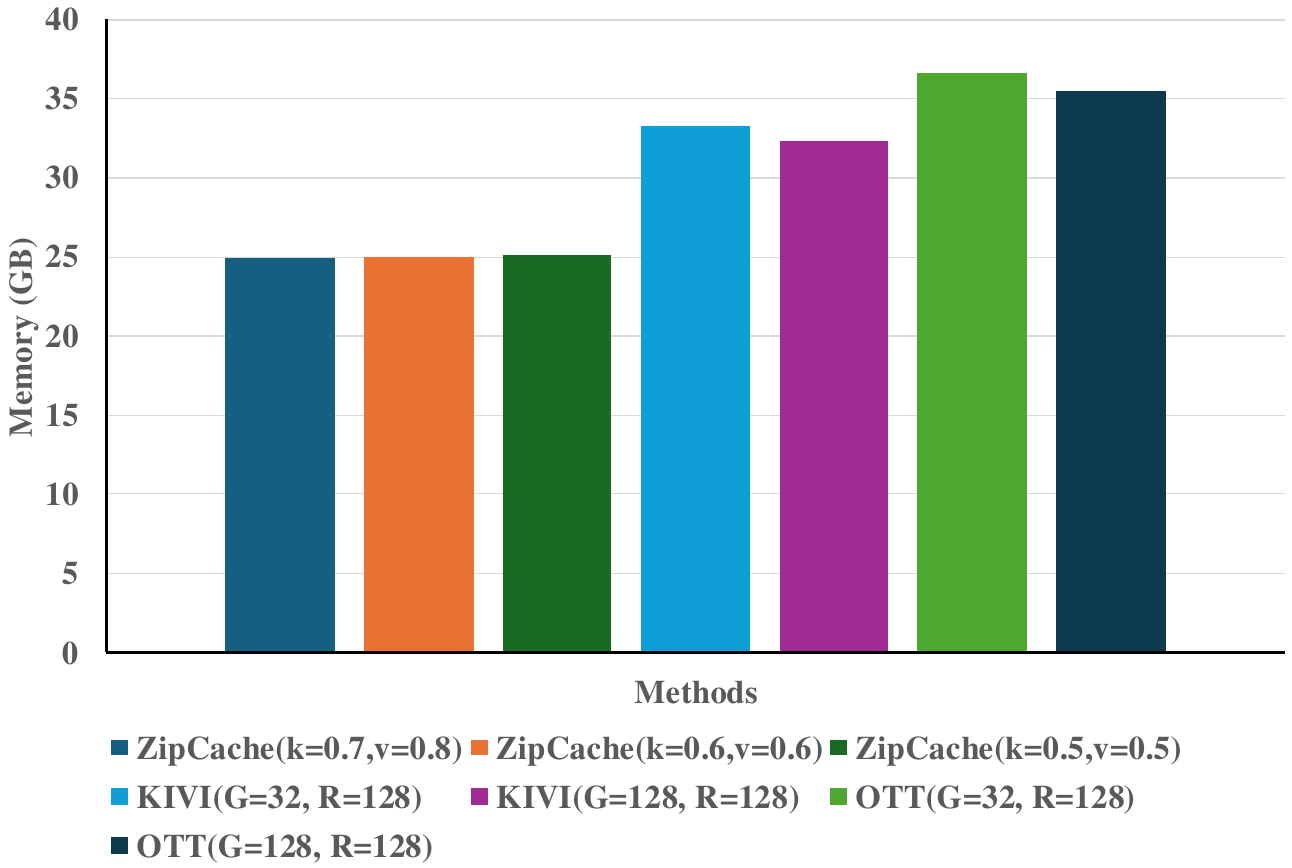}
  \end{subfigure}
  \caption{Throughput (left) and memory usage (right) of different methods under LLaMA-2-7B-chat-hf, input length=64, output length=384, batch size=128 in NVIDIA A100 40G.}
  \label{fig:baselines_memory_speed}
\end{figure*}

\begin{figure*}[htbp]
  \centering
  \includegraphics[width=0.9\linewidth]{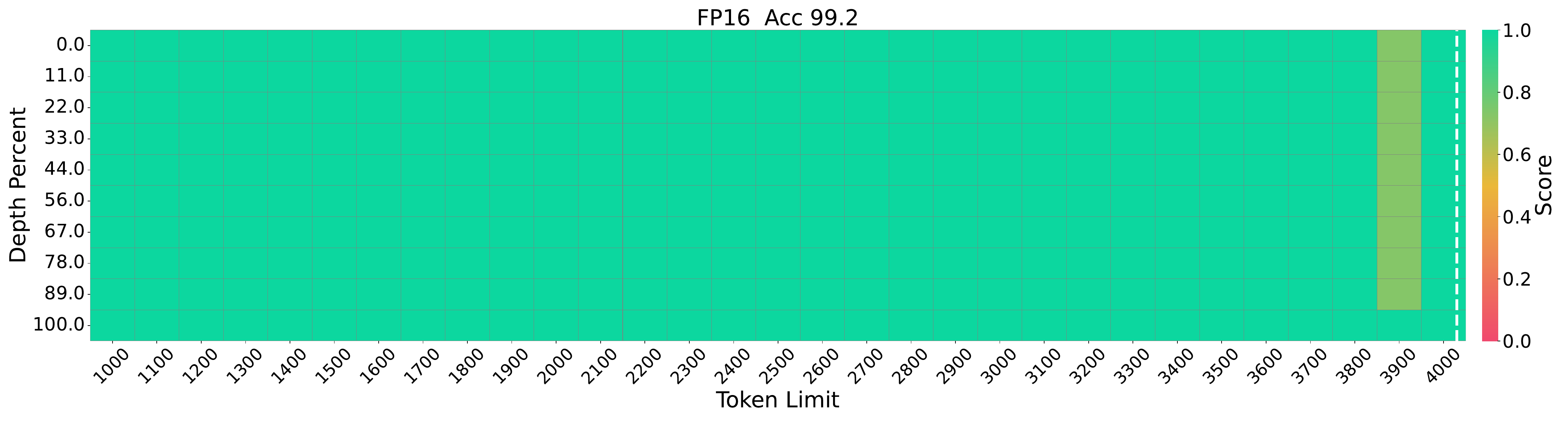}
  \includegraphics[width=0.9\linewidth]{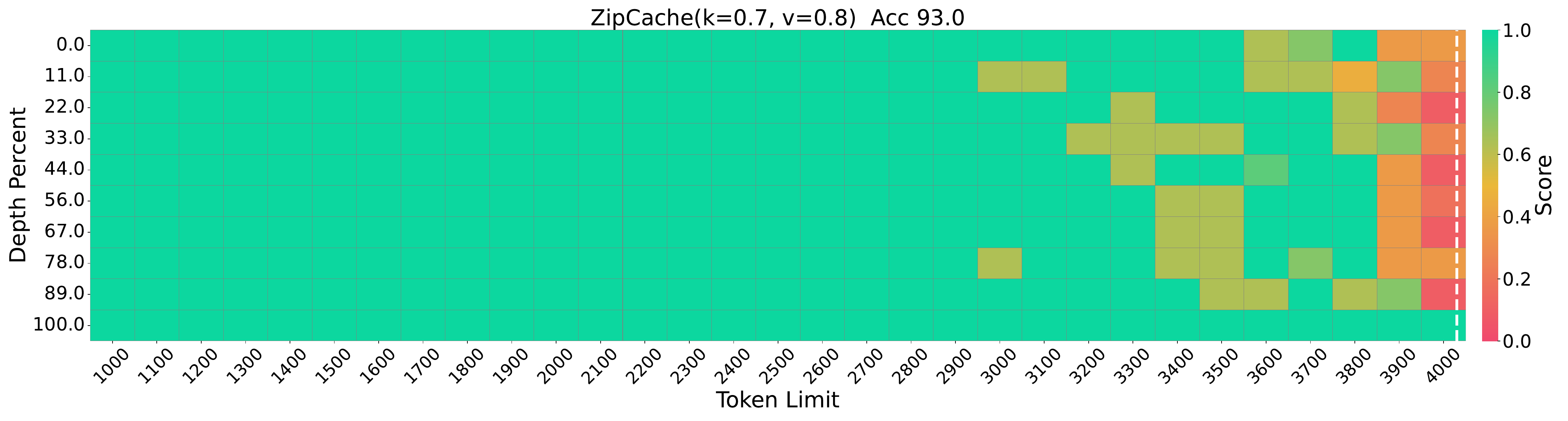}

  \includegraphics[width=0.9\linewidth]{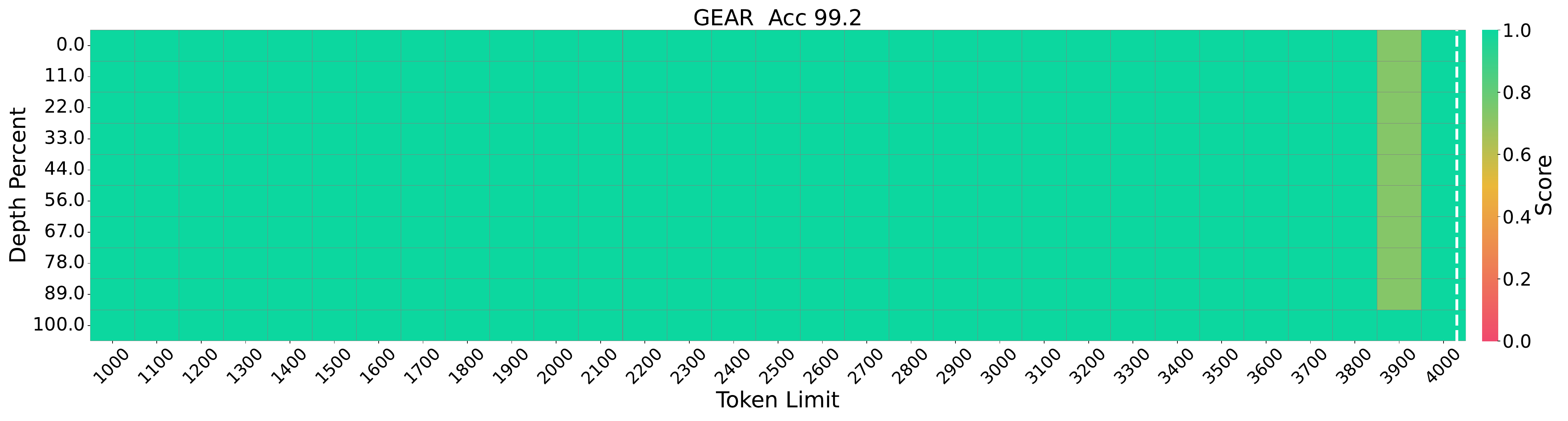}
  \includegraphics[width=0.9\linewidth]{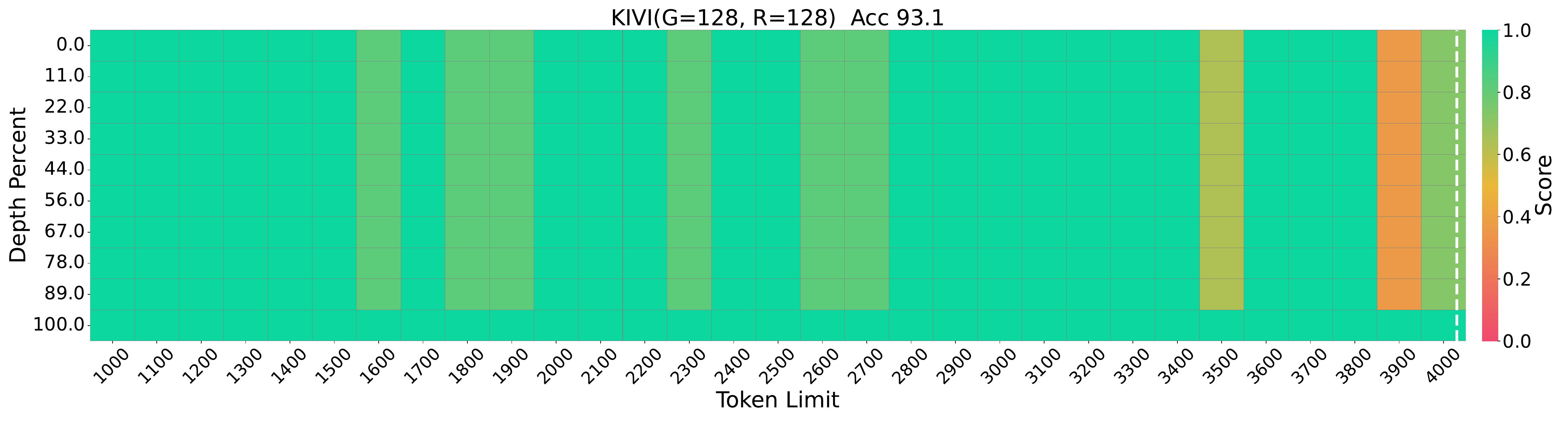}

  \includegraphics[width=0.9\linewidth]{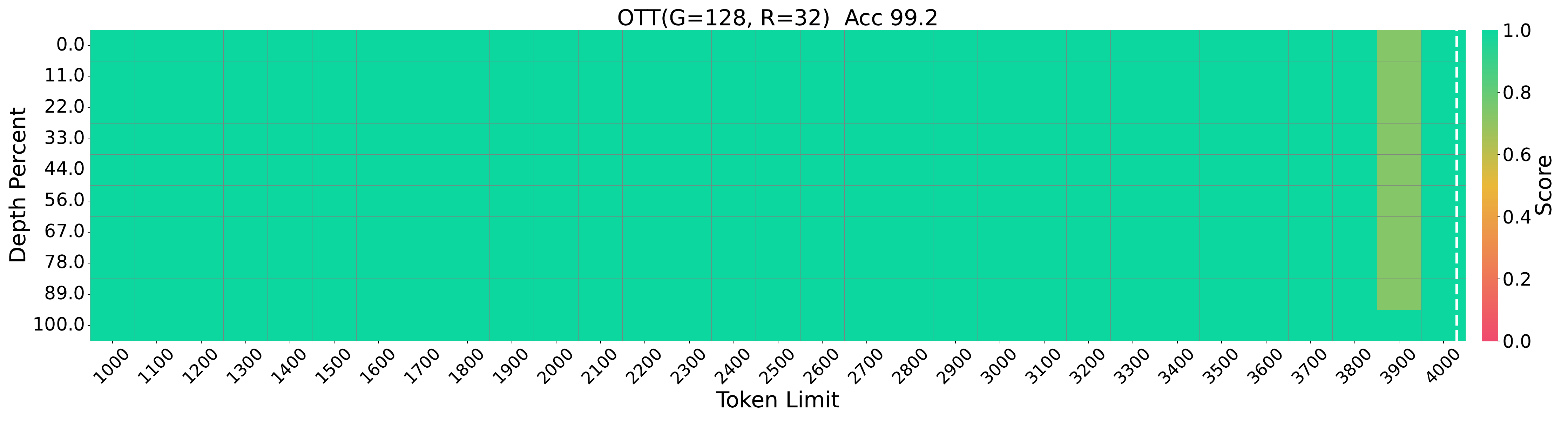}
  \caption{Results of Needle-in-a-Haystack on LLaMA-2-7B-chat-hf with 4k context size. The vertical axis of the table represents the depth percentage, and the horizontal axis represents the token length.}
    \label{fig:needle_res1}
\end{figure*}

\begin{figure*}[htbp]
  \centering
  \includegraphics[width=0.9\linewidth]{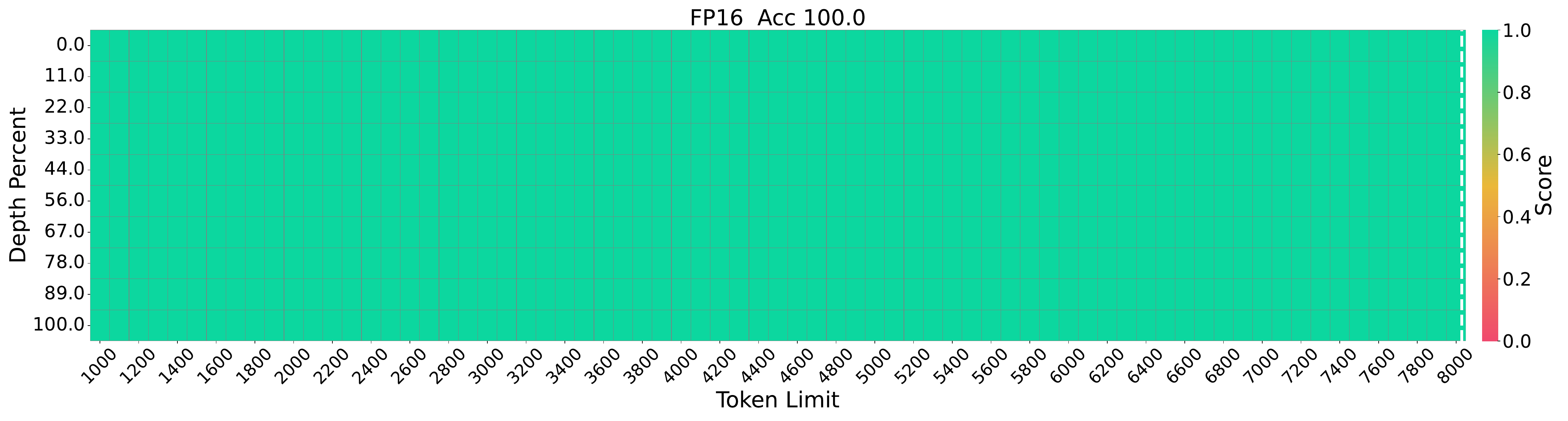}
  \includegraphics[width=0.9\linewidth]{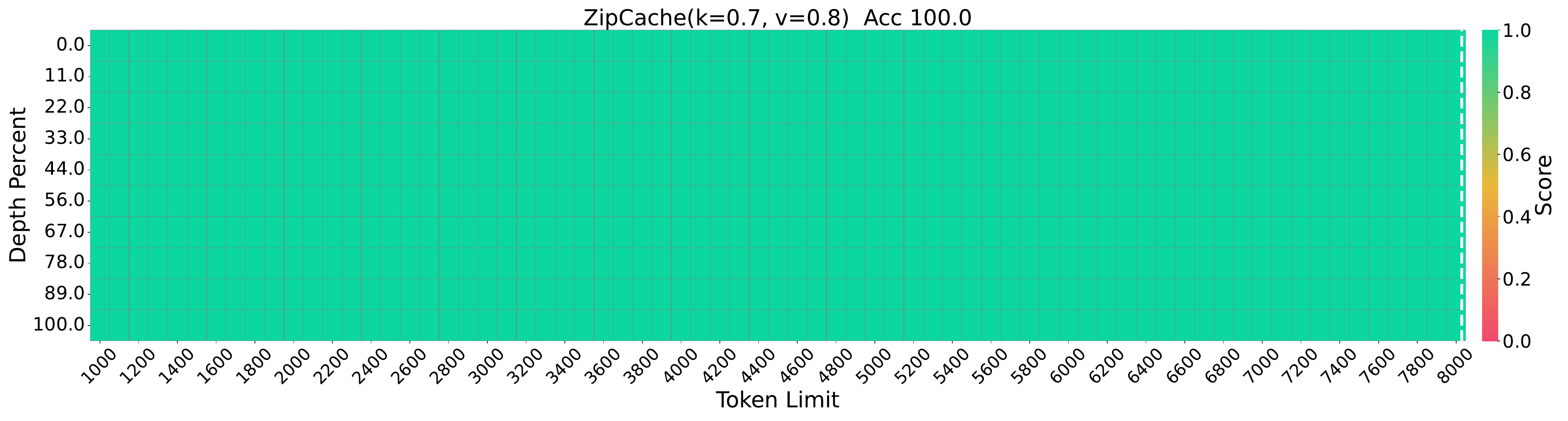}

  \includegraphics[width=0.9\linewidth]{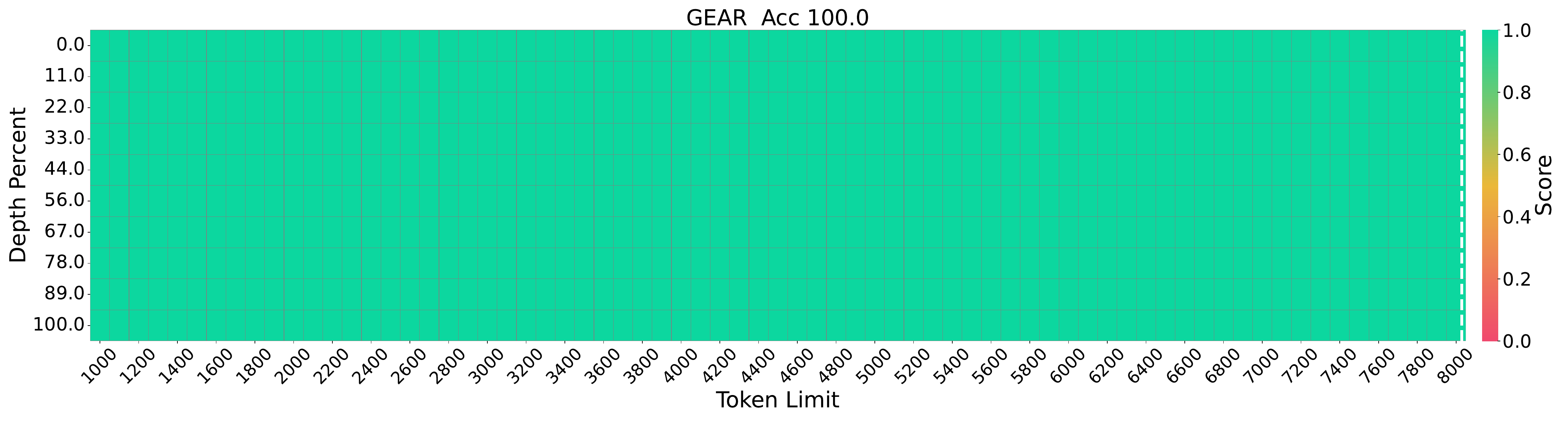}
  \includegraphics[width=0.9\linewidth]{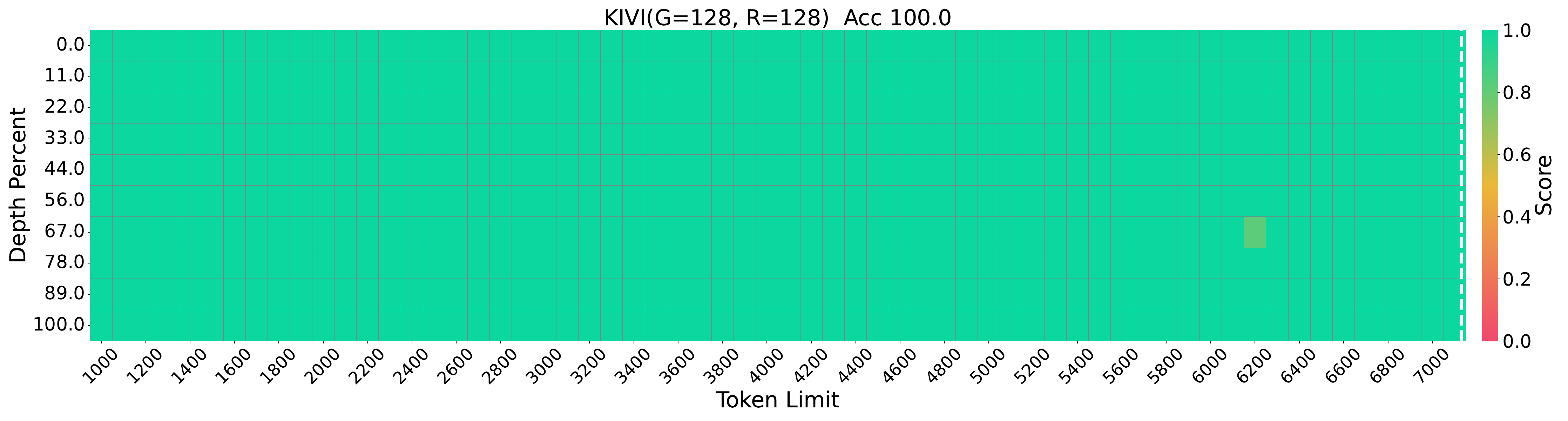}

  \includegraphics[width=0.9\linewidth]{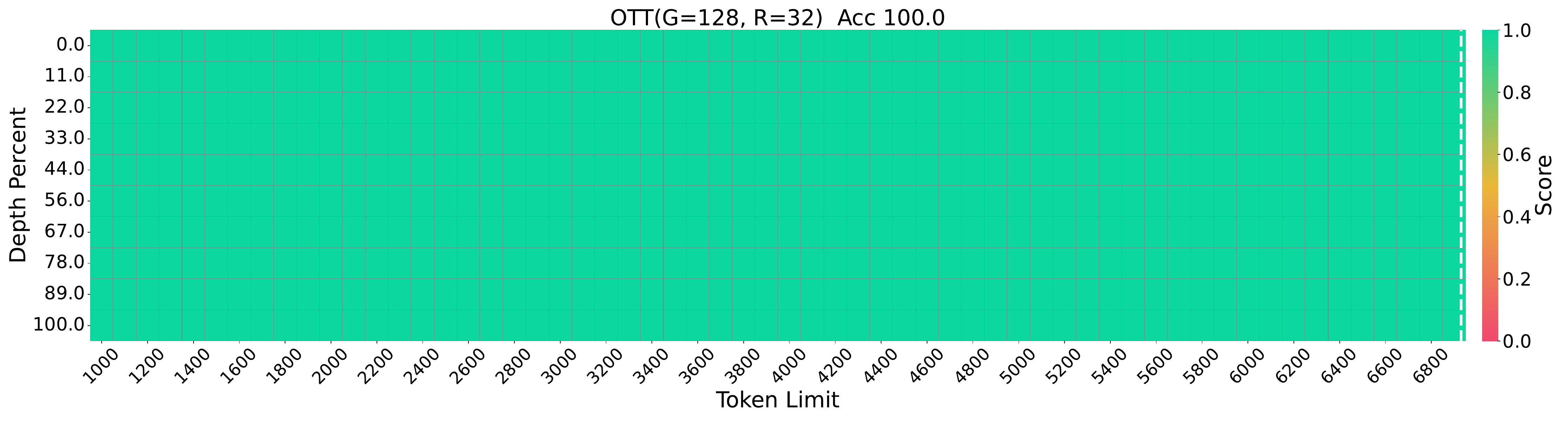}
  \caption{Results of Needle-in-a-Haystack on LLaMA-3-8B-Instruct with 8k context size. The vertical axis of the table represents the depth percentage, and the horizontal axis represents the token length.}
    \label{fig:needle_res2}
\end{figure*}

To fully demonstrate the memory compression and throughput of different methods, we conduct additional experiments on memory usage and throughput.
We use LLaMA-2-7B-chat-hf with an input length of 64, output length of 384, and batch size of 128 to carry out our experiments on an NVIDIA A100 40GB GPU.
We record the throughput and memory peak for each method.
The results are shown in Figure \ref{fig:baselines_memory_speed}.
We omit Gear because its codebase only supports fake compression, making it impossible to measure its actual memory usage and throughput.
Although Gear supports some true compression, it does not handle outliers in its true compression, which is inconsistent with the settings used in our experiments above.
From the figure, we can conclude that OTT has slightly higher throughput than KIVI, likely because its handling of residual tokens is simpler than that of KIVI. Additionally, both OTT and KIVI show significantly higher throughput than ZipCache.
In terms of memory usage, OTT consumes slightly more memory than KIVI, primarily because it needs to store more tokens. This difference may become more pronounced as the batch size increases.
On the other hand, ZipCache uses the least GPU memory, indicating that it has a higher compression ratio.
\begin{table*}[htbp]
\centering
\resizebox{\textwidth}{!}{
\begin{tabular}{|l|c|c|c|c|c|c|c|c|c|c|}
\toprule
\textbf{Dataset} & \textbf{0\%-10\%} & \textbf{10\%-20\%} & \textbf{20\%-30\%} & \textbf{30\%-40\%} & \textbf{40\%-50\%} & \textbf{50\%-60\%} & \textbf{60\%-70\%} & \textbf{70\%-80\%} & \textbf{80\%-90\%} & \textbf{90\%-100\%} \\ \hline
qasper & 0.20 & 0.35 & 0.18 & 0.02 & 0.62 & 8.53 & 29.88 & 37.29 & 19.44 & 3.49 \\ 
triviaqa & 0.20 & 0.36 & 0.16 & 0.01 & 0.31 & 6.42 & 27.36 & 38.78 & 22.23 & 4.17 \\ 
trec & 0.21 & 0.30 & 0.07 & 0.01 & 0.13 & 1.90 & 20.80 & 46.38 & 24.42 & 5.79 \\ 
samsum & 0.20 & 0.41 & 0.18 & 0.01 & 0.18 & 4.33 & 25.58 & 38.52 & 25.36 & 5.24 \\ 
lcc & 0.20 & 0.31 & 0.18 & 0.05 & 0.63 & 7.03 & 25.32 & 36.57 & 24.85 & 4.85 \\ 
repobench-p & 0.20 & 0.34 & 0.14 & 0.04 & 0.58 & 7.00 & 27.65 & 38.84 & 21.22 & 3.99 \\ 
multi\_news & 0.20 & 0.37 & 0.18 & 0.01 & 0.45 & 7.56 & 28.99 & 36.67 & 20.78 & 4.79 \\ 
multifieldqa\_en & 0.20 & 0.37 & 0.21 & 0.02 & 0.71 & 8.17 & 28.25 & 37.12 & 20.93 & 4.01 \\ 
hotpotqa & 0.20 & 0.38 & 0.16 & 0.01 & 0.39 & 7.62 & 30.12 & 40.05 & 18.62 & 2.45 \\ 
2wikimqa & 0.20 & 0.38 & 0.17 & 0.02 & 0.33 & 6.68 & 28.88 & 40.49 & 20.06 & 2.81 \\ 
gov\_report & 0.20 & 0.35 & 0.29 & 0.02 & 0.48 & 6.71 & 27.46 & 39.02 & 21.63 & 3.84 \\ 
passage\_count & 0.20 & 0.36 & 0.16 & 0.01 & 0.44 & 6.62 & 28.51 & 38.87 & 20.91 & 3.92 \\ 
passage\_retrieval\_en & 0.20 & 0.38 & 0.17 & 0.01 & 0.60 & 7.17 & 28.39 & 38.01 & 21.11 & 3.97 \\ 
\bottomrule
\end{tabular}
}
\caption{Statistical analysis of outlier distribution.}
\label{tab:statistical}
\end{table*}

\begin{table*}[htbp]
\centering
\resizebox{\textwidth}{!}{
\begin{tabular}{l|cccccccccccc}

\specialrule{1pt}{0pt}{2pt}
\multirow{4}{*}{Method}  & \multicolumn{3}{c}{Single NIAH} & \multicolumn{3}{c}{Multi-key NIAH} & \multirow{4}{*}{\rotatebox[origin=c]{30}{MQ-NIAH}} & \multirow{4}{*}{\rotatebox[origin=c]{30}{MV-NIAH}} & \multirow{4}{*}{\rotatebox[origin=c]{30}{CWE}} & \multirow{4}{*}{\rotatebox[origin=c]{30}{FWE}} & \multirow{4}{*}{\rotatebox[origin=c]{30}{VT}} & \multirow{4}{*}{Avg.} \\
\cmidrule(lr){2-4}\cmidrule(lr){5-7}
& \rotatebox[origin=c]{30}{S-NIAH-1} & \rotatebox[origin=c]{30}{S-NIAH-2} & \rotatebox[origin=c]{30}{S-NIAH-3} & \rotatebox[origin=c]{30}{MK-NIAH-1} & \rotatebox[origin=c]{30}{MK-NIAH-2} & \rotatebox[origin=c]{30}{MK-NIAH-3} &  \\

\arrayrulecolor{black}\midrule
\multicolumn{13}{c}{LLaMA-2-7B-chat-hf} \\
\arrayrulecolor{black!20}\midrule
FP16 & 100.00 & 92.80 & 90.00 & 84.00 & 67.40 & 52.80 & 76.85 & 80.45 & 83.72 & 80.67 & 92.12 & 81.89 \\
GEAR & 47.40 & 42.20 & 42.40 & 43.60 & 36.00 & 19.60 & 41.05 & 43.10 & 53.86 & 75.13 & 49.60 & 44.90 \\
ZipCache(k=0.7,v=0.8) & 39.60 & 27.80 & 10.40 & 23.60 & 5.00 & 0.00 & 24.50 & 18.95 & 52.02 & 68.73 & 44.48 & 28.64 \\
ZipCache(k=0.6,v=0.6) & 42.60 & 33.40 & 17.40 & 31.40 & 6.40 & 0.00 & 30.95 & 28.15 & 52.98 & 70.40 & 44.04 & 32.52 \\
ZipCache(k=0.5,v=0.5) & 43.00 & 35.80 & 21.60 & 36.20 & 9.20 & 0.20 & 34.40 & 31.70 & 54.68 & 71.33 & 44.36 & 34.77 \\
KIVI(G=128,R=128) & 47.00 & 39.40 & 26.80 & 41.40 & 13.20 & 0.20 & 37.05 & 38.40 & 52.76 & 71.27 & 45.84 & 37.57 \\
KIVI(G=32,R=128) & 46.20 & 42.20 & 36.60 & 41.40 & 25.20 & 3.80 & 41.55 & 42.20 & 58.94 & 73.27 & 47.40 & 41.71 \\
Ours(G=128,R=32) & 46.80 & 39.60 & 28.60 & 41.40 & 18.40 & 0.20 & 38.75 & 39.75 & 52.04 & 72.27 & 46.60 & 38.58 \\
Ours(G=32,R=128) & 46.40 & 42.00 & 38.40 & 43.00 & 27.20 & 5.40 & 41.35 & 41.95 & 55.16 & 74.13 & 48.76 & 42.16 \\
												
\arrayrulecolor{black!20}\midrule
\multicolumn{13}{c}{LLaMA-3-8B-Instruct} \\
\arrayrulecolor{black!20}\midrule
FP16 & 100.00 & 98.20 & 97.00 & 99.20 & 91.60 & 95.80 & 99.75 & 97.45 & 97.82 & 82.27 & 98.28 & 96.12 \\
GEAR & 100.00 & 98.20 & 97.00 & 99.20 & 91.80 & 87.80 & 99.70 & 96.80 & 97.82 & 81.53 & 98.24 & 95.28 \\
ZipCache(k=0.7,v=0.8) & 99.80 & 97.00 & 77.80 & 91.60 & 69.20 & 12.40 & 94.45 & 94.40 & 96.82 & 81.73 & 95.64 & 82.80 \\
ZipCache(k=0.6,v=0.6) & 99.60 & 97.80 & 82.00 & 93.80 & 73.00 & 19.80 & 97.45 & 96.05 & 96.88 & 82.47 & 96.16 & 85.00 \\
ZipCache(k=0.5,v=0.5) & 99.80 & 97.40 & 83.40 & 96.80 & 78.20 & 31.20 & 98.65 & 97.55 & 96.98 & 82.53 & 96.60 & 87.19 \\
KIVI(G=128,R=128) & 96.00 & 97.80 & 88.60 & 96.00 & 75.20 & 11.00 & 95.30 & 94.90 & 86.96 & 80.27 & 88.64 & 82.79 \\
KIVI(G=32,R=128) & 100.00 & 97.40 & 95.40 & 97.80 & 87.60 & 62.40 & 98.80 & 98.60 & 95.26 & 82.20 & 96.20 & 91.97 \\
Ours(G=128,R=32) & 99.80 & 97.20 & 92.20 & 96.20 & 79.40 & 27.60 & 96.80 & 95.45 & 95.38 & 80.47 & 94.76 & 86.84 \\
Ours(G=32,R=128) & 100.00 & 97.60 & 96.20 & 97.80 & 86.00 & 69.60 & 98.90 & 97.85 & 97.18 & 82.20 & 95.80 & 92.64 \\
												
\arrayrulecolor{black}\bottomrule
\end{tabular}
}
\caption{Performance comparison of different methods on RULER for LLaMA-2-7B-chat-hf and LLaMA-3-8B-Instruct. Bold text represents the best performance.}
\label{table:ruler}
\vspace{-3mm}
\end{table*}

\begin{table*}[htbp]

\centering

\resizebox{\textwidth}{!}{
\begin{tabular}{l|cccccccccccccc}
\specialrule{1pt}{0pt}{2pt}
\multirow{5}{*}{Method}  & \multicolumn{2}{c}{Single-Document QA} & \multicolumn{2}{c}{Multi-Document QA}& \multicolumn{2}{c}{Summarization}& \multicolumn{3}{c}{Few-shot Learning}& \multicolumn{2}{c}{Synthetic} & \multicolumn{2}{c}{Code} & \multirow{5}{*}{Avg.} \\
\cmidrule(lr){2-3}\cmidrule(lr){4-5}\cmidrule(lr){6-7}\cmidrule(lr){8-10}\cmidrule(lr){11-12}\cmidrule(lr){13-14}
& \rotatebox[origin=c]{30}{MF-en} & \rotatebox[origin=c]{30}{Qasper} & \rotatebox[origin=c]{30}{HotpotQA} & \rotatebox[origin=c]{30}{2WikiMQA} & \rotatebox[origin=c]{30}{GovReport} & \rotatebox[origin=c]{30}{MultiNews} & \rotatebox[origin=c]{30}{TREC} & \rotatebox[origin=c]{30}{TriviaQA} & \rotatebox[origin=c]{30}{SAMSum} & \rotatebox[origin=c]{30}{PCount} & \rotatebox[origin=c]{30}{PRe} & \rotatebox[origin=c]{30}{Lcc} & \rotatebox[origin=c]{30}{RB-P} & \\
\cmidrule(lr){2-14}
&18409&3619&9151&4887&8734&2113&5177&8209&6258&11141&9289&1235&4206& \\

\arrayrulecolor{black}\midrule
\multicolumn{15}{c}{LLaMA-2-7B-chat-hf} \\
\arrayrulecolor{black!20}\midrule
FP16 & 20.04 & 85.39 & 59.67 & 39.28 & 59.59 & 48.04 & 23.02 & 34.34 & 35.19 & 31.94 & 25.08 & 6.33 & 15.33 & 37.17 \\
GEAR & 19.36 & 85.58 & 59.67 & 38.34 & 58.03 & 46.44 & 20.32 & 34.94 & 34.24 & 32.06 & 21.41 & 6.33 & 14.67 & 36.26 \\
ZipCache(k=0.7,v=0.8)& 19.20 & 84.03 & 59.33 & 38.14 & 53.26 & 45.81 & 18.48 & 28.05 & 33.23 & 30.12 & 17.31 & 6.48 & 12.67 & 34.32 \\
ZipCache(k=0.6,v=0.6) & 19.28 & 84.31 & 59.00 & 39.41 & 56.54 & 46.28 & 19.47 & 29.98 & 33.83 & 31.13 & 18.58 & 7.44 & 14.67 & 35.38 \\
ZipCache(k=0.5,v=0.5) & 19.54 & 85.02 & 59.33 & 39.50 & 55.75 & 45.41 & 20.40 & 29.80 & 33.94 & 31.09 & 19.80 & 6.33 & 14.67 & 35.43 \\
KIVI(G=128,R=128) & 20.43 & 85.16 & 59.67 & 37.70 & 58.73 & 47.24 & 19.82 & 31.03 & 34.65 & 30.38 & 19.97 & 6.33 & 11.67 & 35.60 \\
KIVI(G=32,R=128) & 19.92 & 84.92 & 59.67 & 38.08 & 58.04 & 47.05 & 22.03 & 31.75 & 34.77 & 31.89 & 22.64 & 7.00 & 14.00 & 36.29 \\
Ours(G=128,R=32) & 19.95 & 85.00 & 59.67 & 39.10 & 59.44 & 48.51 & 20.81 & 34.12 & 34.98 & 31.87 & 21.56 & 6.33 & 11.00 & 36.33 \\
Ours(G=32,R=128) & 21.34 & 84.94 & 59.67 & 39.04 & 59.48 & 47.64 & 22.09 & 32.38 & 34.64 & 32.43 & 24.43 & 7.33 & 14.33 & \textbf{36.90} \\

\arrayrulecolor{black}\midrule
\multicolumn{15}{c}{LLaMA-3-8B-Instruct} \\
\arrayrulecolor{black!20}\midrule
FP16 & 37.54 & 89.85 & 69.67 & 40.50 & 56.58 & 51.01 & 25.58 & 40.56 & 49.81 & 34.93 & 31.04 & 12.94 & 83.67 & 47.98 \\
GEAR & 37.55 & 89.85 & 69.67 & 40.02 & 56.42 & 50.47 & 25.52 & 40.11 & 49.80 & 34.93 & 30.93 & 12.61 & 83.33 & \textbf{47.79} \\
ZipCache(k=0.7,v=0.8) & 36.91 & 89.98 & 69.33 & 40.71 & 42.30 & 44.84 & 23.98 & 41.88 & 49.01 & 33.87 & 28.13 & 14.35 & 82.00 & 45.95 \\
ZipCache(k=0.6,v=0.6) & 36.61 & 90.14 & 69.33 & 40.58 & 45.82 & 44.96 & 24.70 & 39.32 & 50.05 & 33.46 & 29.31 & 13.60 & 84.33 & 46.32 \\
ZipCache(k=0.5,v=0.5) & 35.86 & 89.86 & 69.00 & 39.97 & 46.41 & 44.64 & 25.36 & 40.36 & 49.75 & 33.69 & 30.12 & 12.53 & 83.33 & 46.22 \\
KIVI(G=128,R=128) & 34.88 & 89.57 & 69.33 & 40.09 & 44.42 & 45.54 & 24.78 & 39.19 & 49.65 & 34.19 & 28.43 & 11.51 & 82.00 & 45.66 \\
KIVI(G=32,R=128) & 37.27 & 89.88 & 70.00 & 40.46 & 47.29 & 45.20 & 25.34 & 41.29 & 49.87 & 35.05 & 30.38 & 12.67 & 83.67 & 46.80 \\
Ours(G=128,R=32) & 36.75 & 89.74 & 69.67 & 40.39 & 52.37 & 48.82 & 24.94 & 41.57 & 50.37 & 35.32 & 30.74 & 11.44 & 83.33 & 47.34 \\
Ours(G=32,R=128) & 36.71 & 90.36 & 70.00 & 40.67 & 52.65 & 47.76 & 25.34 & 39.60 & 50.36 & 35.13 & 31.16 & 12.33 & 84.33 & 47.42 \\

\arrayrulecolor{black}\bottomrule
\end{tabular}
}
\caption{Performance comparison of OTT with GEAR, ZipCache, KIVI and FP16 on LongBench for LLaMA-3-8B-Instruct and LLaMA-2-7B-chat-hf. OTT generally achieves improvements over previous KV cache compression methods across various LLMs. Bold text represents the best performance.}
\label{table:longbench}
\vspace{-3mm}
\end{table*}

\begin{table*}[htbp]
\centering
\renewcommand{\arraystretch}{1.4} 
\setlength{\tabcolsep}{3.0pt} 
\resizebox{2\columnwidth}{!}{%
\begin{tabular}{l|cccccccccccccc}
\toprule

\multirow{2}{*}{Method}  & \multicolumn{3}{c}{ruler\_recall} & \multicolumn{2}{c}{substring\_exact\_match}& \multicolumn{1}{c}{NDCG@10}& \multicolumn{1}{c}{str\_em}& \multicolumn{2}{c}{citation\_rec} & \multicolumn{1}{c}{qampari\_rec\_top5} & \multicolumn{2}{c}{citation\_prec} & \multirow{2}{*}{Avg.} \\
\cmidrule(lr){2-4}\cmidrule(lr){5-6}\cmidrule(lr){7-7}\cmidrule(lr){8-8}\cmidrule(lr){9-10}\cmidrule(lr){11-11} \cmidrule(lr){12-13} 
& \rotatebox[origin=c]{30}{niah\_mk\_2} & \rotatebox[origin=c]{30}{niah\_mk\_3} & \rotatebox[origin=c]{30}{niah\_mv} & \rotatebox[origin=c]{30}{json\_kv} & \rotatebox[origin=c]{30}{hotpotqa} & \rotatebox[origin=c]{30}{rerank\_psg} & \rotatebox[origin=c]{30}{alce\_asqa} & \rotatebox[origin=c]{30}{alce\_qampari} & \rotatebox[origin=c]{30}{alce\_asqa} & \rotatebox[origin=c]{30}{alce\_qampari} & \rotatebox[origin=c]{30}{alce\_asqa} & \rotatebox[origin=c]{30}{alce\_qampari} \\
\midrule
full & 100.00 & 100.00 & 99.75 & 98.00 & 61.00 & 55.27 & 41.37 & 8.47 & 6.03 & 17.80 & 12.88 & 9.46 & 50.84 \\
Gear & 100.00 & 92.00 & 99.75 & 92.00 & 60.67 & 54.01 & 42.07 & 7.20 & 6.19 & 18.00 & 10.85 & 8.22 & \textbf{49.25} \\
KIVI(G=32,R=128) & 99.00 & 85.00 & 99.50 & 77.00 & 59.67 & 41.52 & 41.00 & 5.83 & 7.79 & 14.80 & 7.72 & 7.81 & 45.55 \\
ZipCache(k=0.5,v=0.5) & 98.00 & 51.00 & 99.25 & 45.00 & 59.00 & 44.88 & 35.47 & 5.49 & 7.28 & 6.80 & 7.44 & 4.95 & 38.71 \\
Ours(G=32,R=128) & 99.00 & 87.00 & 99.75 & 80.00 & 61.00 & 42.92 & 44.50 & 9.52 & 6.51 & 17.40 & 12.01 & 8.33 & 47.33 \\
\bottomrule
\end{tabular}
}
\caption{Performance comparison of OTT with GEAR, ZipCache, KIVI and FP16 on HELMET for LLaMA-3-8B-Instruct. OTT generally achieves improvements over previous KV cache compression methods across various LLMs. Bold text represents the best performance.} 
\label{tab:helmet} 
\end{table*}
\begin{table*}[htbp]
\centering
\small
\renewcommand{\arraystretch}{1.4} 
\setlength{\tabcolsep}{3.0pt} 
\begin{tabular}{l|ccccccc}
\toprule
\textbf{Method} & \textbf{Easy} & \textbf{Hard} & \textbf{Short} & \textbf{Medium} & \textbf{Long} & \textbf{Overall} \\
\midrule
FP16 & 27.6 & 27.0 & 25.6 & 25.6 & 33.3 & 27.2 \\
GEAR & 27.6 & 27.0 & 25.6 & 25.6 & 33.3 & \textbf{27.2} \\
KIVI (G=32, R=128) & 26.6 & 22.2 & 23.9 & 21.9 & 27.8 & 23.9 \\
ZipCache (k=0.5, v=0.5) & 26.6 & 25.4 & 26.1 & 23.3 & 30.6 & 25.8 \\
Ours (G=32, R=128) & 25.5 & 26.7 & 26.7 & 22.8 & 32.4 & 26.2 \\
\bottomrule
\end{tabular}
\caption{Performance comparison of OTT with GEAR, ZipCache, KIVI and FP16 on LongBench\_v2 for LLaMA-3-8B-Instruct. OTT generally achieves improvements over previous KV cache compression methods across various LLMs. Bold text represents the best performance.}
\label{tab:longbench2} 
\end{table*}
\begin{table*}[htbp]
\centering
\resizebox{\textwidth}{!}{
\begin{tabular}{l|cccccccccccc}

\specialrule{1pt}{0pt}{2pt}
\multirow{4}{*}{Method}  & \multicolumn{3}{c}{Single NIAH} & \multicolumn{3}{c}{Multi-key NIAH} & \multirow{4}{*}{\rotatebox[origin=c]{30}{MQ-NIAH}} & \multirow{4}{*}{\rotatebox[origin=c]{30}{MV-NIAH}} & \multirow{4}{*}{\rotatebox[origin=c]{30}{CWE}} & \multirow{4}{*}{\rotatebox[origin=c]{30}{FWE}} & \multirow{4}{*}{\rotatebox[origin=c]{30}{VT}} & \multirow{4}{*}{Avg.} \\
\cmidrule(lr){2-4}\cmidrule(lr){5-7}
& \rotatebox[origin=c]{30}{S-NIAH-1} & \rotatebox[origin=c]{30}{S-NIAH-2} & \rotatebox[origin=c]{30}{S-NIAH-3} & \rotatebox[origin=c]{30}{MK-NIAH-1} & \rotatebox[origin=c]{30}{MK-NIAH-2} & \rotatebox[origin=c]{30}{MK-NIAH-3} &  \\

\arrayrulecolor{black}\midrule
\multicolumn{13}{c}{max\_length=64k} \\
\arrayrulecolor{black!20}\midrule
FP16 & 100 & 99.4 & 100 & 99 & 99.8 & 99.4 & 98.85 & 95.8 & 8.42 & 76 & 97.96 & 88.60 \\
GEAR & 100 & 99.4 & 100 & 99 & 99.6 & 90.2 & 94.4 & 96 & 8.62 & 76.53 & 97.96 & \textbf{87.43} \\
KIVI(G=32,R=128) & 99.8 & 99.2 & 97.8 & 95.6 & 97.2 & 78.4 & 94.5 & 93 & 8.26 & 78.73 & 92.24 & 84.98 \\
Ours(G=32,R=128) & 99.8 & 99.2 & 99.2 & 96 & 97.6 & 82 & 97.5 & 94.5 & 9.28 & 78.53 & 93.92 & 86.14 \\
												
\arrayrulecolor{black!20}\midrule
\multicolumn{13}{c}{max\_length=128k} \\
\arrayrulecolor{black!20}\midrule
FP16 & 100 & 94 & 100 & 93 & 100 & 100 & 98.75 & 96.25 & 0.3 & 82.67 & 96.6 & 87.42 \\
GEAR & 100 & 94 & 98 & 93 & 100 & 87 & 98.5 & 96.25 & 0.3 & 81.67 & 96.6 & \textbf{85.94} \\
KIVI(G=32,R=128) & 100 & 92 & 99 & 91 & 95 & 69 & 93 & 87.5 & 0.4 & 80.33 & 89.4 & 81.51 \\
Ours(G=32,R=128) & 100 & 94 & 99 & 92 & 95 & 73 & 96.25 & 90.25 & 0.4 & 82.67 & 88.8 & 82.85 \\
												
\arrayrulecolor{black}\bottomrule
\end{tabular}
}
\caption{Performance comparison of Llama-3-8B-ProLong-512k-Instruct with longer context lengths (64k, 128k).}
\label{table:longer_ruler}
\vspace{-3mm}
\end{table*}

\clearpage
\end{document}